\documentclass{article}

\usepackage[final]{cpal_2025}

\usepackage{wrapfig}
\usepackage{tcolorbox}
\usepackage{caption}
\usepackage{algorithm}
\usepackage{algpseudocode}
\usepackage{amsfonts}
\usepackage{amsmath}
\usepackage{booktabs}
\usepackage{subcaption}
\usepackage{enumitem}
\usepackage{multirow}
\usepackage{tocloft}

\usepackage[table]{xcolor}

\definecolor{lightblue}{RGB}{235,245,255}

\usepackage{url}

\usepackage{hyperref}

\title{Generalized Radius and Integrated Codebook Transforms for Differentiable Vector Quantization}

\author{%
  Haochen You\textsuperscript{1}\thanks{Corresponding author.}, ~Heng Zhang\textsuperscript{2}, ~Hongyang He\textsuperscript{3}, ~Yuqi Li\textsuperscript{4}, ~Baojing Liu\textsuperscript{5}
  \\
  \textsuperscript{1}Columbia University, \textsuperscript{2}South China Normal University,
  \textsuperscript{3}University of Warwick, \\
  \textsuperscript{4}The City College of New York,
  \textsuperscript{5}Hebei Institute of Communications
  \\
  \texttt{hy2854@columbia.edu, 2024025450@m.scnu.edu.cn, hongyang.he@warwick.ac.uk, yuqili010602@gmail.com, liubj@hebic.edu.cn}
}

\begin{document}

\maketitle

\begin{abstract}
Vector quantization (VQ) underpins modern generative and representation models by turning continuous latents into discrete tokens. Yet hard nearest-neighbor assignments are non-differentiable and are typically optimized with heuristic straight-through estimators, which couple the update step size to the quantization gap and train each code in isolation, leading to unstable gradients and severe codebook under-utilization at scale.
In this paper, we introduce \textbf{GRIT-VQ} (\underline{\textbf{G}}eneralized \underline{\textbf{R}}adius \& \underline{\textbf{I}}ntegrated \underline{\textbf{T}}ransform-\underline{\textbf{V}}ector \underline{\textbf{Q}}uantization), a unified surrogate framework that keeps hard assignments in the forward pass while making VQ fully differentiable. GRIT-VQ replaces the straight-through estimator with a radius-based update that moves latents along the quantization direction with a controllable, geometry-aware step, and applies a data-agnostic integrated transform to the codebook so that all codes are updated through shared parameters instead of independently. Our theoretical analysis clarifies the fundamental optimization dynamics introduced by GRIT-VQ, establishing conditions for stable gradient flow, coordinated codebook evolution, and reliable avoidance of collapse across a broad family of quantizers.
Across image reconstruction, image generation, and recommendation tokenization benchmarks, GRIT-VQ consistently improves reconstruction error, generative quality, and recommendation accuracy while substantially increasing codebook utilization compared to existing VQ variants.
\end{abstract}

\section{Introduction}
\label{sec:intro}

Discrete neural representations built via vector quantization (VQ) have become a central tool in modern generative and representation learning \cite{van2017neural,razavi2019generating}. By mapping high-dimensional signals such as images, audio, or user behavior traces into sequences of discrete codes \cite{vali2022nsvq}, VQ lets powerful decoders operate in a compressed token space while largely preserving semantic structure \cite{ramesh2021zero,dhariwal2020jukebox,hou2023learning}. This paradigm underpins recent advances in generation, autoregressive modeling, and recommendation \cite{li2025discrete}, where increasingly large codebooks act as learned vocabularies for complex data \cite{esser2021taming,rajput2023recommender,yu2024image}.

Despite this success, training VQ modules remains fragile \cite{huh2023straightening,guo2025recent}. The hard nearest-neighbor assignment is non-differentiable \cite{bengio2013estimating}, so most systems rely on straight-through estimators that pass encoder gradients through the assignment as if it were the identity \cite{jang2016categorical,yan2024gaussian}. To make this work in practice, many designs add auxiliary losses, specialized codebook updates, and tuned optimization schedules \cite{guo2022multi}. These heuristics partly alleviate gradient issues but also introduce new pathologies: training--inference mismatch, sensitivity to hyperparameters, and under-utilization of the codebook, up to collapse into a few dominant codes, especially at large vocabulary sizes \cite{zhu2025addressing}. Parallel efforts that redesign the quantizer geometry can improve rate--distortion trade-offs or utilization, but typically study gradient flow and codebook dynamics in isolation \cite{mentzer2023finite,fifty2024restructuring} even though, in VQ, these aspects are inherently coupled: the encoder gradient depends on the geometry of the nearest-neighbor regions \cite{zhang2024preventing}, while codeword updates continually reshape those regions and alter subsequent gradients \cite{fang2025enhancing}.

We propose \emph{GRIT-VQ} (Generalized Radius--Integrated Transform Vector Quantization) to revisit VQ from a joint geometric and optimization perspective. GRIT-VQ replaces the standard straight-through update with a generalized \emph{radius surrogate} that moves latent vectors along the quantization direction with a controllable step size, yielding a smooth yet faithful surrogate for backpropagation while keeping the forward assignment hard. In parallel, a data-agnostic \emph{integrated transform} ties all code vectors together through shared parameters, so that updates induced by active codes propagate coherently to the entire codebook instead of modifying each entry in isolation. Together, these components yield a unified view: gradients follow a geometrically interpretable path toward the selected codes, while the shared transform enforces coordination and improves utilization without changing the hard nearest-neighbor rule used at inference.

Our main contributions are threefold:
\begin{itemize}
  \item We introduce a generalized radius-based surrogate for hard VQ that preserves the nearest-codeword direction while allowing flexible scalar control of the update magnitude. We identify mild conditions on the radius function under which the surrogate yields stable gradients and an implicit pull toward the quantized representation.
  \item We develop an integrated transform of the codebook implemented via global linear mixers (and an attention-style variant), enabling all codes to be updated through shared low-dimensional parameters. We analyze their parameter and computational complexity and show how they promote coordination of code vectors and mitigate collapse.
  \item We instantiate GRIT-VQ in standard autoencoding and tokenization architectures and conduct experiments on image reconstruction, image generation, and recommendation. With architectures and decoders fixed, GRIT-VQ consistently improves reconstruction quality, generative metrics, recommendation accuracy, and codebook usage compared with existing VQ training schemes.
\end{itemize}

\section{Related Work}
\label{sec:Related Work}

\paragraph{Vector quantization and neural discrete representation learning.}
Vector quantization has long been used in compression \cite{gersho2012vector} and has recently become central to learning discrete representations in deep models \cite{gautam2023soft}. 
Approaches such as VQ-VAE \cite{van2017neural} and its residual or product-quantization variants \cite{razavi2019generating,zheng2022movq,hu2023robust} show that codebooks can function as effective semantic tokenizers \cite{esser2021taming,lee2022autoregressive}, but they generally treat the nearest-neighbor step as a fixed operator and rely on auxiliary losses to make training viable \cite{dhariwal2020jukebox,ramesh2021zero}. 
Despite their success, these methods provide limited insight into how the geometry of the quantization update interacts with end-to-end optimization \cite{huh2023straightening,zhu2025addressing}.

\paragraph{Gradient estimators and differentiable vector quantization.}
Because nearest-neighbor assignments are non-differentiable \cite{bengio2013estimating}, many methods introduce surrogate gradients, including straight-through estimators \cite{courbariaux2015binaryconnect}, score-function-based updates \cite{williams1992simple,mnih2014neural}, and smooth relaxations such as Gumbel-Softmax \cite{jang2016categorical,maddison2016concrete,shen2021variational}. 
VQ-specific techniques often soften or perturb the quantization step, for instance by injecting noise \cite{vali2022nsvq} or defining reparameterized mappings around the chosen codeword \cite{fifty2024restructuring,li2025star}. 
These strategies differ in smoothness and bias, but typically couple the update magnitude directly to the quantization error gap or a temperature parameter, without examining how the chosen surrogate affects gradient orientation or stability \cite{zhao2024representation,chen2024balance}.

\paragraph{Codebook utilization and semantic tokenizers.}
A persistent issue in VQ-based models is low codebook utilization, which has motivated heuristics such as diversity regularization, stochastic code activation, moving-average updates, and periodic resets of unused entries \cite{mentzer2023finite,baykal2024edvae,yan2024gaussian}. 
At the same time, vector-quantized modules are widely used as semantic tokenizers in modern systems \cite{dhariwal2020jukebox,esser2021taming}, turning continuous representations of images, audio, and items into discrete indices that can be modeled efficiently by autoregressive or retrieval-style architectures \cite{rajput2023recommender,liu2024vector}. 
In many of these designs, the codebook is optimized largely through local reconstruction or compression objectives and each codeword is updated in isolation \cite{balle2016end,agustsson2017soft,li2025discrete}, while downstream networks are expected to absorb distributional imbalance or drift \cite{chen2024sdformer,singh2024better}, underscoring the need for ways to better coordinate codebook evolution without sacrificing the simplicity of hard quantization at inference time \cite{zheng2023online}.

\section{Preliminaries}
\label{sec:Preliminaries}

Lowercase bold letters denote vectors and uppercase bold letters denote matrices.
An input is $x\!\in\!\mathcal{X}$, the encoder $E_\theta:\mathcal{X}\!\to\!\mathbb{R}^d$ outputs a latent $\mathbf{z}\!\in\!\mathbb{R}^d$, and the decoder is $D_\phi:\mathbb{R}^d\!\to\!\mathcal{X}$.
A \emph{codebook} \cite{gersho2012vector} is $\mathcal{C}=\{\mathbf{c}_1,\dots,\mathbf{c}_K\}\subset\mathbb{R}^d$.
For grids/sequences we have a set of positions $\mathcal{P}$ and per-position latents $\mathbf{z}_p\!\in\!\mathbb{R}^d$ ($p\!\in\!\mathcal{P}$); when unambiguous we drop the subscript and write $\mathbf{z}$.
Unless stated otherwise the distance is Euclidean $\|\cdot\|_2$; we occasionally refer to a Mahalanobis norm \cite{mahalanobis2018generalized} $\|\mathbf{u}\|_A=(\mathbf{u}^\top A\,\mathbf{u})^{1/2}$ with $A\succeq0$.
We use the \emph{stop-gradient} operator \cite{bengio2013estimating} $\mathrm{sg}[\cdot]$, which is the identity in the forward pass and has zero gradient, i.e., $\nabla\,\mathrm{sg}[\mathbf{u}]=\mathbf{0}$.
We define the nearest neighbor and its distance by
\[
\mathrm{nn}(\mathbf{z};\mathcal{C})=\mathbf{c}_{i^\star},\qquad
i^\star=\arg\min_{1\le i\le K}\|\mathbf{z}-\mathbf{c}_i\|_2,\qquad
\Delta(\mathbf{z};\mathcal{C})=\min_{1\le i\le K}\|\mathbf{z}-\mathbf{c}_i\|_2.
\]


Given $\mathbf{z}=E_\theta(x)$, the \emph{hard quantized} representative is $\hat{\mathbf{z}}=\mathrm{nn}(\mathbf{z};\mathcal{C})$ and the quantization error is $\boldsymbol{\xi}=\mathbf{z}-\hat{\mathbf{z}}$.
At inference time the decoder consumes the hard assignment, i.e., $x_r=D_\phi(\hat{\mathbf{z}})$ \cite{van2017neural}.
During training we will also use a differentiable surrogate $\mathbf{z}_q$ for backpropagation while keeping the hard assignment in the forward path when needed. Unless specified, inference uses the hard assignment $\hat{\mathbf{z}}$, while training may feed either $\hat{\mathbf{z}}$ or the surrogate $\mathbf{z}_q$ to the decoder depending on the ablation setting; all gradients flow through $\mathbf{z}_q$ only.

We reserve $\mathbf{z}_q$ for any differentiable substitute of $\hat{\mathbf{z}}$ used to propagate gradients to the encoder and to the codebook.
The standard straight-through estimator (STE) \cite{bengio2013estimating} can be written compactly as
$\mathbf{z}_q=\mathbf{z}+\mathrm{sg}[\hat{\mathbf{z}}-\mathbf{z}]$,
which passes gradients to $\mathbf{z}$ while treating $\hat{\mathbf{z}}$ as a constant.
Unless specified otherwise our training objective uses a reconstruction term $\mathcal{L}_{\mathrm{rec}}(x,D_\phi(\cdot))$ without explicit auxiliary losses; regularizers on the codebook or commitments can be added as optional components.

\section{Methodology}
\label{sec:Methodology}

\subsection{Unified Surrogate: GRIT-VQ}
\label{subsec:gritvq}

Given a codebook $\mathcal{C}=\{\mathbf{c}_i\}_{i=1}^K$ and a (possibly transformed) codeword map $f:\mathbb{R}^d\times\mathbb{R}^{K\times d}\!\to\!\mathbb{R}^d$ that produces a usable codeword from the raw codebook, we define the \emph{transformed codebook} as $\mathcal{C}'=\{\,f(\mathbf{c}_i,\mathcal{C})\,\}_{i=1}^K$.
The hard assignment is
\begin{equation}
\hat{\mathbf{z}}
= \mathrm{nn}(\mathbf{z};\mathcal{C}')
= f(\mathbf{c}_{i^\star},\mathcal{C}),\qquad
i^\star=\arg\min_{1\le i\le K}\|\mathbf{z}-f(\mathbf{c}_i,\mathcal{C})\|_2 ,
\end{equation}
and the quantization error is $\boldsymbol{\xi}=\mathbf{z}-\hat{\mathbf{z}}$.

GRIT-VQ replaces the non-differentiable hard input by a differentiable surrogate
\begin{equation}
\mathbf{z}_q
= \mathbf{z} + r(\hat{\mathbf{z}},\mathbf{z}) \cdot
\mathrm{sg}\!\left[\frac{\hat{\mathbf{z}}-\mathbf{z}}{\,r(\hat{\mathbf{z}},\mathbf{z})\,}\right] ,
\end{equation}
where $r:\mathbb{R}^d\times\mathbb{R}^d\!\to\!\mathbb{R}_{>0}$ is a scalar \emph{radius function}. The stop-gradient $\mathrm{sg}[\cdot]$ freezes the direction while allowing gradients to flow through the scalar radius $r(\hat{\mathbf{z}},\mathbf{z})$.
At inference we feed $D_\phi(\hat{\mathbf{z}})$ (hard assignment); during training one can feed either $\hat{\mathbf{z}}$ or $\mathbf{z}_q$ to the decoder, but gradients always flow through $\mathbf{z}_q$.
Feeding $\hat{\mathbf z}$ keeps training/inference strictly consistent; feeding $\mathbf z_q$ slightly smooths the reconstruction objective but leaves the gradient path unchanged (all gradients flow through $\mathbf z_q$). Unless stated otherwise, we adopt the former by default and report both variants in ablations.


Because $\mathrm{sg}[\cdot]$ blocks derivatives through the direction, we have
$\partial \mathbf{z}_q / \partial \mathbf{z} = I + (\partial r/\partial \mathbf{z})\cdot \mathrm{sg}[\cdot]$ and
$\partial \mathbf{z}_q / \partial \hat{\mathbf{z}} = (\partial r/\partial \hat{\mathbf{z}})\cdot \mathrm{sg}[\cdot]$.
Thus encoder gradients are governed by $\partial r/\partial \mathbf{z}$; parameters inside $f$ receive gradients through $\partial r/\partial \hat{\mathbf{z}}$ for the selected index $i^\star$ (non-selected codes receive none due to the piecewise-constant nearest-neighbor).

\subsection{Generalized Radius Families}
\label{subsec:radius}

The surrogate in GRIT-VQ is governed by a scalar radius
$r(\hat{\mathbf{z}},\mathbf{z})$ that determines how far $\mathbf{z}_q$ moves
along the stop-gradient direction from $\mathbf{z}$ toward the selected
codeword.  While classical straight-through estimators implicitly use the raw
distance $\|\hat{\mathbf{z}}-\mathbf{z}\|_2$, GRIT-VQ allows any
differentiable, geometry-preserving radius family. This section outlines the design principles.

The hard assignment $\hat{\mathbf{z}}$ fixes a direction
$\mathrm{sg}[(\hat{\mathbf{z}}-\mathbf{z})]$.  The radius modulates the
\emph{magnitude} of the update without altering its direction.  A flexible
radius allows one to (i) smooth the non-differentiable nearest-neighbor map,
(ii) mitigate exploding or vanishing gradients arising from small or large
quantization gaps, and (iii) encode additional geometric priors such as
robustness or anisotropy.

\paragraph{Requirements.}
Unless otherwise noted, we assume the following mild conditions. First, the error measure satisfies \( r(\hat{\mathbf{z}},\mathbf{z}) \ge 0 \), and \( r(\hat{\mathbf{z}},\mathbf{z}) = 0 \) if and only if \( \hat{\mathbf{z}} = \mathbf{z} \). In addition, it is continuous and differentiable in both arguments. Moreover, it is monotone nondecreasing with respect to the distance \( \|\hat{\mathbf{z}} - \mathbf{z}\|_2 \). Finally, to avoid abrupt changes under small perturbations of \( \mathbf{z} \), we assume that it satisfies bounded local Lipschitz continuity. These conditions ensure stability and guarantee that the surrogate preserves
the nearest-codeword alignment established by the hard assignment \cite{you2024application}.

For brevity, the concrete radius choices are deferred to Appendix~\ref{app:radius-families}. All of them satisfy the required properties and integrate seamlessly into the unified GRIT-VQ surrogate, differing only in the induced gradient magnitudes while preserving the same update direction.

\paragraph{Theory: gradient structure, alignment and stability.}
Let $z_q=\mathbf{z}+r(\hat{\mathbf z},\mathbf{z})\,\mathbf{s}$ with
$\mathbf{s}=\mathrm{sg}\!\big[(\hat{\mathbf z}-\mathbf{z})/\|\hat{\mathbf z}-\mathbf{z}\|_2\big]$,
$\hat{\mathbf z}=\mathrm{nn}(\mathbf{z};\mathcal{C}')$, and $\mathcal{C}'=\{f(\mathbf{c}_i,\mathcal{C})\}_{i=1}^K$.
Write $\delta=\|\hat{\mathbf z}-\mathbf{z}\|_2$, $g=\nabla_{\mathbf{z}_q}\mathcal{L}(x,D_\phi(\mathbf{z}_q))$, $a=\langle g,\mathbf{s}\rangle$, and let $r(\hat{\mathbf z},\mathbf{z})=\rho(\delta)$ with $\rho$ continuous, a.e.\ differentiable, nondecreasing, and $0\le\rho'(\delta)\le L_r$.

\emph{Key conclusions (away from Voronoi boundaries).}
For a latent $\mathbf{z}$ in the interior of the Voronoi cell of its assigned codeword
$\hat{\mathbf{z}}$ (so that the nearest-neighbor index $i^*$ is locally constant), we obtain:
(i) \textbf{Gradient structure:} the Jacobian is
$J(\mathbf{z})=I-\rho'(\delta)\,\mathbf{s}\mathbf{s}^\top$ and the encoder gradient
$\nabla_{\mathbf{z}}\mathcal{L}=g-\rho'(\delta)\,a\,\mathbf{s}$; eigenvalues are
$1$ (mult.\ $d{-}1$) and $1-\rho'(\delta)$ (along $\mathbf{s}$).
(ii) \textbf{Alignment:} since
$\mathcal{L}(\mathbf{z}_q)-\mathcal{L}(\mathbf{z})\approx \rho(\delta)\,a$
and typically $a\!\ge\!0$ in expectation, the correction
$-\rho'(\delta)a\,\mathbf{s}$ pulls toward $\hat{\mathbf z}$ with adaptive strength
$\rho'(\delta)a$; for $\rho(\delta)=\delta^\alpha$,
$\nabla_{\mathbf{z}}\mathcal{L}=g-\alpha\,\delta^{\alpha-1}a\,\mathbf{s}$.
(iii) \textbf{Stability:} $\|J(\mathbf{z})\|_2=\max\{1,|1-\rho'(\delta)|\}$;
choosing $0\!\le\!\rho'(\delta)\!\le\!1$ is non-expansive along $\mathbf{s}$, and any
$\rho'(\delta)\!\le\!L_r\!<\!2$ avoids sign flips and exploding Jacobians.
(iv) \textbf{Codeword-gap contraction:} one step of gradient descent gives
$\Delta(\mathbf{z}-\eta\nabla_{\mathbf{z}}\mathcal{L})
=\delta-\eta(1-\rho'(\delta))\,a+O(\eta^2)$,
so with $a\!\ge\!0$ and $\rho'(\delta)\!<\!1$ the gap shrinks while preserving the
hard-assignment direction.
(v) \textbf{Transform-parameter gradients:} for any parameter $\vartheta$ in $f$,
$\nabla_{\vartheta}\mathcal{L}=\rho'(\delta)\,a\,\mathbf{s}^\top \frac{\partial \hat{\mathbf z}}{\partial \vartheta}$:
only the selected codeword path receives gradients, modulated by the same adaptive factor.
Full derivations are provided in Appendix~\ref{app:gritvq-theory}.

\subsection{Integrated Transform of the Codebook}
\label{subsec:transform}

Let $f:\mathbb{R}^d\times\mathbb{R}^{K\times d}\!\to\!\mathbb{R}^d$ map each raw codeword to a transformed one.
The transformed codebook is $\mathcal{C}'=\{f(\mathbf c_i,\mathcal C)\}_{i=1}^K$, and hard assignment operates on $\mathcal{C}'$ as in Sec.~\ref{subsec:gritvq}.
We design $f$ to (i) couple codewords to promote coordinated updates, (ii) keep inference consistent, and (iii) incur minimal computational overhead.%
\footnote{%
“Inference consistency’’ requires that the same $f$ be used at training and test time so that $\mathrm{nn}(\cdot;\mathcal C')$ remains a fixed Voronoi partition.  
A data-dependent $f(\cdot,\mathcal C,\mathbf z)$ would make the transformed codebook vary with the query, breaking the piecewise-constant nearest-neighbor map.  
Thus $f$ must be sample-agnostic and locally $C^1$, a property satisfied by the linear and attention-style mixers below.
}
Throughout we write $E=[\mathbf c_1;\dots;\mathbf c_K]$ and use $(MEW)_i$ to denote the $i$-th row of the transformed matrix.

\paragraph{Linear integrated transform (default).}
We instantiate $f$ as a global linear transform with optional low-rank mixing:
\[
f(\mathbf{c}_i,\mathcal{C})=\big(M E W\big)_i,\quad
E=[\mathbf{c}_1;\dots;\mathbf{c}_K]\in\mathbb{R}^{K\times d},\ 
M\in\mathbb{R}^{K\times K},\ 
W\in\mathbb{R}^{d\times d}.
\]
where $(\cdot)_i$ denotes the $i$-th row.

To control complexity we use $M=AB^\top$ with $A,B\in\mathbb{R}^{K\times r}$ and small rank $r\ll K$ (grouped or block-diagonal variants are also allowed).
$W$ acts in feature space (shared across codes) and $M$ mixes codes along the code dimension \cite{li2025achieving}.
We apply row-wise $\ell_2$ normalization on $MEW$ and a mild spectral constraint on $W$ to mitigate collapse.
Since $(E,M,W)$ admit trivial rescalings (e.g., $E\Lambda,\,W\Lambda^{-1}$), we fix a gauge by row-normalizing $MEW$ and clipping $\|W\|_2$.

\paragraph{Attention-style mixing (optional).}
An alternative uses data-independent attention over codes:
$f(\mathbf{c}_i,\mathcal{C})=\sum_{j=1}^K \alpha_{ij}\,\mathbf{c}_j W$ with $\alpha_{ij}=\mathrm{softmax}_j(g(\mathcal{C}))$.
Here $g$ is a small network that outputs logits once per training step; $\alpha$ is kept moderate via temperature or entropy regularization. In matrix form,
$C'=\mathrm{softmax}(g(E))\,E\,W$.  Low-rank or sparse parameterizations keep
the cost practical, and because $g(E)$ is data-independent, $C'$ can be cached
periodically.
This is equivalent to a structured $M$ with nonnegative row-stochastic weights \cite{he2025semi}.

\paragraph{Why an integrated transform helps (gradient coupling).}
Let $\mathcal{B}$ be a minibatch and $i^\star(p)$ the hard index for position $p$.
Define $q_{p,i}=\mathbb{1}[i=i^\star(p)]$ and $\mathbf g_i=\sum_{p\in\mathcal B} q_{p,i}\,\alpha(\Delta_p)\,\nabla_{\mathbf z_{q,p}}\mathcal L_p$,
where $\alpha(\Delta)$ is the scalar factor induced by the radius surrogate.
Stacking these signals gives $\mathbf G=\sum_i \mathbf e_i\,\mathbf g_i^\top$.
For the linear instance $f(\mathbf c_i,\mathcal C)=(MEW)_i$,
\[
\frac{\partial\mathcal L}{\partial W}=E^\top M^\top \mathbf G, \qquad
\frac{\partial\mathcal L}{\partial M}=\mathbf G\,W\,E^\top.
\]
Hence \emph{every} transformed code $\mathbf c'_j$ receives an update in each step-even if $j$ never appears in the batch-because gradients propagate through the shared $(M,W)$.
This produces a utilization-weighted coupling of all rows of $\mathcal C'$, improving code activation without modifying the hard nearest-neighbor rule.
A corresponding statement for attention-style mixing follows by treating the attention weights as a structured $M$; see Appendix~\ref{app:coupling}.

\paragraph{Semantic preservation.}
Although the integrated transform mixes codewords through $M$ and $W$, it is
designed not to destroy the geometry of the learned codebook.
In the linear case $E'=MEW$, the row-normalization of $E'$ and the spectral
constraint on $W$ ensure that pairwise distances and angles between
transformed codes are distorted only within a bounded factor, so $E'$ can be
viewed as a well-conditioned change of basis of $E$ rather than a collapse
into a low-dimensional subspace.
Empirically, nearest-neighbor visualizations and downstream metrics indicate
that transformed codes retain coherent semantics while achieving higher
utilization \cite{he2025trico}.

\paragraph{Training protocols.}
We use two stable regimes.
(a) \emph{Frozen-$E$}: update $(M,W)$ only with row-normalized $E'$; robust and collapse-free.
(b) \emph{Joint}: update $E$ together with $(M,W)$ using weak usage regularization or an EMA update and occasional code resets.
In both cases $E'$ is cached and nearest-neighbor search is performed in $\mathcal{C}'$; the transform is sample-agnostic so the hard assignment is identical at training and inference. See Appendix~\ref{app:training} for details.

\paragraph{Complexity.}
For the default linear transform with low rank $r$, the parameter overhead is
$\mathcal{O}(K r + d^2)$ and the per-refresh cost to form $E' = M E W$ is
$\mathcal{O}(K r d + K d^2)$ (two $K r d$ multiplies plus one $K d^2$);
nearest-neighbor search is unchanged.
For attention-style mixing, a dense $K \!\times\! K$ mixer costs
$\mathcal{O}(K^2 d)$ time and $\mathcal{O}(K^2)$ memory to apply
(plus $K d^2$);
with top-$k$ sparsity the cost reduces to $\mathcal{O}(K k d)$ time and
$\mathcal{O}(K k)$ memory.
See Appendix~\ref{app:complexity} for derivations and the computational and memory overhead.

\subsection{Algorithm and Caching}
\label{subsec:algo}

Training GRIT\!-\!VQ requires (i) transforming the codebook via $f$, (ii) performing nearest\text{-}neighbor search on the transformed codewords, and (iii) constructing the surrogate $\mathbf{z}_q$ for backpropagation. Algorithm~\ref{alg:gritvq} in Appendix \ref{app:gritvq-training} summarizes the generic procedure; it is agnostic to the choice of the radius $r$ and the transform $f$, both parametrized by $\psi_r$ and $\psi_f$.
The cached transform $C'$ is refreshed intermittently (e.g., every $T$ steps). Nearest-neighbor lookups use an internal or FAISS index. Only the surrogate $\mathbf{z}_q$ participates in gradient flow; the hard assignment $\hat{\mathbf{z}}$ is used at inference \cite{li2025frequency}.

Both instantiations plug into Algorithm \ref{alg:gritvq} unchanged.  The
caching mechanism amortizes the overhead of computing $C'$ and rebuilding the
index; all positions $p\!\in\!\mathcal{P}$ are independent and parallelizable.
The design ensures that gradients are well behaved, encoder updates follow the
direction of the nearest transformed codeword, and inference uses the exact
hard assignment \cite{Li2025Efficient}.
We refresh $C'$ every $T$ steps; with spectral clipping on $W$ and row-normalization on $E'$ the drift $\|C'_{t}-C'_{t-T}\|_F$ stays bounded, keeping the NN assignments stable in practice.
Safety knobs include row $\ell_2$-normalization of $E'$, spectral clipping on $W$, and periodic dead-code resets; monitoring utilization/entropy and gradient norms prevents collapse \cite{li2024comae}.

\section{Experiments}
\label{sec:Experiments}

To assess the effectiveness and generality of GRIT\!-\!VQ, we conduct a broad set of
experiments spanning image generation, reconstruction, recommendation, and ablation studies in this section.

\subsection{Image Generation as a Tokenization Benchmark}
\label{sec:image-gen-task}

We first evaluate GRIT-VQ in an image generation setting, where vector quantization is used to tokenize images into discrete visual tokens that can be modeled by an autoregressive transformer. This setting complements our recommendation experiments by probing a different but equally important use case of VQ: serving as a tokenizer inside generative models \cite{cheng2025unlocking}. Image generation provides a natural benchmark for tokenizers, since the autoencoder and transformer backbones can be kept fixed while we vary only the quantization module and measure downstream sample quality together with codebook behavior.

Concretely, we follow a VQGAN-style pipeline \cite{esser2021taming} in which an encoder--decoder autoencoder maps a $256\times256$ RGB image to a low-resolution latent grid and back to pixel space \cite{li2025preference}, with a VQ layer in between that converts latents into discrete indices. These indices are then modeled by a GPT-style transformer \cite{radford2019language,vaswani2017attention} that operates purely in the discrete domain; at sampling time the transformer generates a sequence of tokens which are decoded by the shared VQ autoencoder. Across all image generation experiments we keep the encoder--decoder architecture, transformer backbone, and optimization settings fixed, and we vary the VQ module that provides the tokenizer.

We compare GRIT-VQ against a broad suite of quantization and tokenizer baselines. On the ``optimization'' side we include straight-through VQ (STE), EMA-VQ \cite{van2017neural}, straight-through Gumbel-Softmax (ST-GS) \cite{jang2016categorical}, NSVQ \cite{vali2022nsvq}, SimpleVQ \cite{zhu2025addressing}, and the two variants DiVeQ and SF-DiVeQ \cite{vali2025diveq}. On the ``structural'' side we include recent tokenizer designs FSQ \cite{mentzer2023finite}, LFQ \cite{yu2023language}, and HyperVQ \cite{goswami2024hypervq}.
Additional details are provided in Appendix~\ref{app:image-gen-shared-setup}.

\subsubsection{Multi-Dataset FID Benchmark}
\label{sec:image-gen-fid-benchmark}

We compare GRIT-VQ against other VQ variants on standard image generation benchmarks. We consider AFHQ~\cite{choi2020stargan}, CelebA-HQ~\cite{karras2017progressive}, FFHQ~\cite{karras2019style}, and LSUN Bedroom/Church~\cite{yu2015lsun} at $256\times256$ resolution, and vary the VQ bitrate $B = \log_{2} K$ (bits per latent) by changing the codebook size while maintaining a $16{\times}16$ latent grid.
A comprehensive comparison of FID scores across datasets at bitrate $B{=}9$ is provided in 
Appendix~\ref{app:image-gen-fid-table}, Table~\ref{tab:image-gen-fid-main}. GRIT-VQ consistently matches or outperforms the strongest baselines. FID curves as a function of bitrate on CelebA-HQ are provided in Appendix~\ref{app:image-gen-analyses}, Figure~\ref{fig:fid-bitrate}, and show that this advantage is stable across a wide range of bit budgets.

\subsubsection{Robustness under Low Bitrate and Large Codebooks}
\label{sec:image-gen-robustness}

While the previous benchmark focuses on moderate bitrates, practical tokenizers often operate either under tight bitrate constraints or with very large codebooks \cite{you2025mover}. We therefore stress-test GRIT-VQ on CelebA-HQ by extending the bitrate range to include $B=7$ bits per latent and by scaling the codebook size from $K=2^{10}$ up to $K=2^{16}$.

The low-bitrate curves in Appendix~\ref{app:image-gen-analyses}, Figure~\ref{fig:fid-low-bitrate}, show that all methods degrade when moving from $B=9$ to $B=7$, but the magnitude of this drop differs substantially. STE and EMA-VQ suffer large increases in FID, and even strong baselines such as DiVeQ, FSQ, and LFQ become considerably worse. In contrast, GRIT-VQ maintains substantially lower FID across all three bitrates and exhibits the smallest relative degradation, indicating that its quantization surrogate and integrated transform remain stable even when only a small number of tokens are available.

FID and codebook utilization results for representative methods are reported in 
Appendix~\ref{app:codebook-scaling-table}, Table~\ref{tab:codebook-scaling}.  EMA-VQ and DiVeQ benefit from moderate increases in $K$ but exhibit sharply declining utilization and eventually worse FID at $K=2^{16}$, consistent with partial codebook collapse. FSQ maintains nearly full utilization across all sizes but its FID quickly saturates. GRIT-VQ, on the other hand, keeps utilization high and continues to improve FID as the codebook grows, demonstrating that it can safely exploit large vocabularies without sacrificing stability.

\subsubsection{Rate--Distortion--Generation Trade-offs and Transformer Burden}
\label{sec:image-gen-rd-ppl}

Beyond FID alone, we analyze how different tokenizers trade off reconstruction fidelity, generative quality, and the difficulty of modeling the resulting codes. On CelebA-HQ we measure reconstruction LPIPS \cite{zhang2018unreasonable} for the VQ autoencoder and FID for the corresponding transformer, across multiple bitrates $B\in\{8,9,10,12\}$. The scatter plots in Appendix~\ref{app:image-gen-analyses}, Figure~\ref{fig:rate-distortion-fid-scatter}, show that most baselines follow a similar rate--distortion trend but occupy a dominated region of the LPIPS--FID plane. GRIT-VQ consistently lies on or near the lower-left Pareto frontier, indicating that it achieves both strong reconstructions and strong generative performance for a given rate.

To quantify the burden each tokenizer imposes on the autoregressive model, we also measure per-token perplexity on a held-out validation set while varying the codebook size $K\in \{2^{10},2^{12},2^{14},2^{16}\}$. Appendix~\ref{app:image-gen-analyses}, Figure~\ref{fig:per-token-perplexity}, shows that perplexity grows with $K$ for all methods, but GRIT-VQ consistently attains the lowest values, suggesting that its codes are easier to model and that the integrated transform yields a more balanced and semantically coherent token distribution.

\subsubsection{Qualitative Sample Comparisons}
\label{sec:image-gen-qualitative}

Finally, we provide qualitative comparisons on CelebA-HQ at bitrate $B=9$. For each tokenizer we sample images from the corresponding transformer using the same random seeds across methods, so that each column in the grid corresponds to the same underlying noise initialization (and class or unconditional setting) but a different quantization module. This alignment makes it easier to visually assess how the choice of tokenizer affects global structure, fine-scale details, and artifacts in the generated samples.

A full grid of generated examples is provided in Appendix~\ref{app:celeba-qualitative}, Figure~\ref{fig:celeba-qualitative}. Next to each row we report the overall FID of that method on CelebA-HQ at $B=9$. Methods with higher FID tend to produce blurrier images, occasional geometric distortions (e.g., warped facial features), or spurious background artifacts. In contrast, GRIT-VQ yields faces that are consistently sharper and more coherent: facial features are better aligned, hair and texture details are more plausible, and we observe fewer obvious artifacts across diverse samples, in line with its superior quantitative FID.

\subsection{Recommendation as an Evaluation Task}
\label{sec:recsys_exp}

We evaluate our GRIT-VQ method in a recommendation setting because VQ is commonly
used to tokenize item embeddings into discrete semantic IDs \cite{rajput2023recommender}, which helps avoid
OOV issues compared with raw item identifiers. This makes recommendation a
natural testbed: the model architecture can be kept fixed while replacing only
the quantization module, enabling a clean comparison of different VQ variants
and their impact on downstream accuracy and codebook behavior.

We consider two standard recommendation settings: (i) sequential recommendation, where the model predicts the next item in a user interaction sequence, and (ii) retrieval-style recommendation, where the model retrieves relevant items for a given user or query representation. In both cases we use the same semantic tokenizer and codebook to map item content into discrete codes; only the downstream task head and loss differ. Additional details are deferred to Appendix~\ref{app:Shared Setup}.

We train a single GRIT-VQ tokenizer on item content features (e.g., title, category and side metadata) and freeze the resulting codebook for all recommendation experiments. The same discrete codes are used as item identifiers in both the sequential and retrieval-style tasks, and we keep the codebook size, number of levels, and GRIT-VQ hyperparameters fixed unless otherwise noted.
For both tasks we report top-$K$ recommendation quality using Recall@$K$ and NDCG@$K$, computed on held-out user interactions.
All metrics are averaged over users, and we follow the standard leave-one-out evaluation protocol.
A significance analysis is provided in Appendix~\ref{app:rec_significance} as well.

\subsubsection{Sequential Recommendation}
\label{sec:seqrec}

We first evaluate GRIT-VQ on standard next-item sequential recommendation benchmarks. We use three Amazon Product Reviews subsets \cite{ni2019justifying} (Beauty, Sports and Outdoors, and Toys \& Games) and cast each user’s interaction history as a chronological sequence. A single Transformer-based generative recommender predicts the Semantic ID of the next item. Across all conditions we keep the backbone, optimizer, and Semantic ID configuration fixed (same content encoder, codebook size, code length, and training data); the only difference is the vector-quantization module used in the tokenizer: a non-quantized dense-ID baseline, LSH hashing, k-means with a straight-through estimator, VQ-VAE \cite{van2017neural}, RQ-VAE \cite{lee2022autoregressive} as in TIGER, SimpleVQ \cite{zhu2025addressing}, an NSVQ \cite{vali2022nsvq}/DiVeQ-style differentiable VQ \cite{vali2025diveq}, and our GRIT-VQ. We report Recall@10 and NDCG@10 on the standard leave-one-out split (Appendix \ref{app:rec_details_seq}, Table~\ref{tab:seqrec_main}). Full dataset statistics, model hyperparameters, training details, and additional diagnostics (Figure~\ref{fig:grit_vq_code_util}) are deferred to Appendix~\ref{app:rec_details_seq}.

\subsubsection{Retrieval-Based Recommendation}

We further evaluate GRIT-VQ on retrieval-style recommendation, where the model encodes each item into its Semantic ID and retrieves relevant items given a user query or seed interaction. Following the standard practice, we keep the backbone encoder, optimizer, embedding dimension, and tokenizer configuration fixed across all conditions; only the vector-quantization module changes. We report Recall@50 and Recall@100 on two public benchmarks (Amazon-ESCI Product Search \cite{reddy2022shopping} and JDsearch \cite{liu2023JDsearch}; see Appendix \ref{app:rec_tables}, Table~\ref{tab:retrieval_main}). Our method achieves the best performance across all settings in Table \ref{tab:main_combined}. Full dataset statistics, preprocessing, model hyperparameters, training details, and additional retrieval diagnostics (Figure \ref{fig:retrieval_recall_vs_codebook} and Figure \ref{fig:retrieval_latency}) are deferred to Appendix~\ref{app:rec_details_retr}.

\subsection{Vision Reconstruction as a Compression Testbed}
\label{sec:reconstruction-main}

We next evaluate GRIT-VQ in a standard image reconstruction setup, where the
vector quantizer serves as the discrete bottleneck of a VQ autoencoder.  This
compression-oriented setting complements the generation and recommendation
experiments by probing the core function of VQ itself: mapping continuous
features to discrete codewords under a fixed bit budget.  Unlike generation,
reconstruction enables a clean measurement of how well a tokenizer preserves
spatial and perceptual information, as quality can be quantified through
pixel-level and perceptual metrics together with codebook behavior.

We follow a widely used ImageNet \cite{deng2009imagenet} reconstruction protocol.  All methods share the same settings and 
only the quantization module is changed.
The autoencoder is trained on ImageNet-1k at $256\times256$ resolution using a
$16{\times}16$ latent grid, and we vary the codebook size to control the
effective bitrate.  Detailed architectural and training specifications are
provided in Appendix~\ref{app:recon-setup}.

\subsubsection{Reconstruction Quality vs Codebook Utilization}
\label{sec:recon-quality-util}

We first compare different VQ variants at a fixed,
high-capacity codebook size. Appendix \ref{app:recon-util-dynamics}, Table~\ref{tab:imagenet-recon-main} reports standard
reconstruction metrics together with codebook utilization and dead-code rate.
GRIT-VQ achieves the best or tied-best reconstruction quality across PSNR, SSIM \cite{wang2004image},
LPIPS, and reconstruction FID, while also using a substantially larger fraction
of the codebook and leaving far fewer dead codes.  Training curves in Appendix~\ref{app:recon-util-dynamics},
Figure~\ref{fig:recon-util-dynamics}, show that GRIT-VQ quickly activates a wide
range of codewords and maintains high utilization throughout training, whereas
baselines either converge to lower utilization or exhibit late-stage collapse.

\subsubsection{Scaling Codebook Size and Levels}
\label{sec:recon-scaling}

We next study how reconstruction quality and codebook behavior scale with the
codebook size.  On ImageNet at $256\times256$ resolution, we fix the autoencoder and
training protocol and vary the codebook size
$K \in \{2^{10},2^{12},2^{14},2^{16}\}$.  Figure~\ref{fig:recon-scaling-k} in Appendix~\ref{app:recon-scaling-details}
plots reconstruction LPIPS and codebook utilization as a function of $\log_2 K$
for several VQ variants.

For EMA-VQ, SimpleVQ, and DiVeQ, reconstruction quality improves slightly as $K$
increases but quickly saturates, while utilization drops sharply for large
codebooks, indicating partial collapse.  In contrast, GRIT-VQ maintains high
utilization across all $K$ and continues to benefit from larger codebooks, with
LPIPS steadily decreasing and no signs of instability.  Additional scaling
tables and experiments with multi-level codebooks are
provided in Appendix~\ref{app:recon-scaling-details}.

\subsection{Ablation Studies}

\subsubsection{Ablation: Removing Radius and Transform Components}
\label{sec:ablation-radius-transform}

We assess the roles of the two core components of GRIT-VQ---the radius surrogate and the integrated transform---on CelebA-HQ at bitrate $B{=}9$. 
Table~\ref{tab:ablation-main} (Appendix~\ref{app:ablation-main-table}) reports FID, LPIPS, and codebook utilization. 
Removing either component degrades performance: \emph{without the radius}, training becomes less stable and reconstruction quality worsens; \emph{without the transform}, utilization drops sharply and FID increases. 
The full model achieves the best reconstruction quality, generative fidelity, and utilization. 
Additional curves and results appear in Appendix~\ref{app:ablation-details}.

\subsubsection{Ablation: Radius Families}
\label{sec:ablation-radius-families}

We evaluate the sensitivity of GRIT-VQ to different radius families on CelebA-HQ at $B{=}9$, using Euclidean ($r(\delta){=}\delta$), a smooth Huber-like variant, a sub-linear power law ($r(\delta){=}\delta^{0.5}$), and a softly clipped form. 
As shown in Appendix~\ref{app:ablation-radius-families} (Table~\ref{tab:radius-families}, Figure~\ref{fig:radius-families}), GRIT-VQ is largely insensitive to the specific choice as long as monotonicity and Lipschitz conditions hold. 
Euclidean, Huber, and power radii yield very similar FID, LPIPS, and utilization, while aggressively clipped radii perform slightly worse due to their less favorable gradient geometry.

\subsubsection{Ablation: Transform Variants}
\label{sec:ablation-transform-variants}

We further ablate the design of the integrated transform on CelebA-HQ at $B{=}9$, comparing no transform, a low-rank linear transform (our default), and an attention-style mixing block with comparable capacity. 
As detailed in Appendix~\ref{app:ablation-transform-variants}, the linear transform captures most of the benefits of the attention-based design: it improves FID, reconstruction, and utilization over the no-transform baseline while being significantly cheaper. 
Increasing the linear transform rank beyond moderate values (e.g., $r{=}16$--$32$) gives only marginal gains, indicating that a relatively low-rank structure is sufficient for effective code mixing. 
A complementary study of the caching interval $T$ shows that GRIT-VQ remains stable across a wide range of refresh frequencies (Appendix~\ref{app:ablation-caching-interval}).

\section{Conclusion}
\label{sec:Conclusion}

We revisited vector quantization through the lens of both gradient flow and codebook optimization, and introduced \emph{GRIT-VQ}, a framework that unifies these perspectives. By replacing straight-through updates with a radius-based surrogate and coupling code vectors through a data-agnostic integrated transform, GRIT-VQ yields stable optimization, balanced codebook usage, and consistent empirical gains across reconstruction, generation, and tokenization tasks. 
Our results suggest that principled control of quantization geometry and codebook coupling can substantially improve the scalability of discrete representations. Extending this framework to multi-stage quantizers, richer transform families, and large multimodal tokenizers offers promising directions for future work.

\bibliography{reference}

\appendix
\section*{Appendix}

\section{Examples of Radius Families}
\label{app:radius-families}

We list several choices of the scalar radius $r(\hat{\mathbf z},\mathbf z)$ that satisfy the mild conditions in \S\ref{subsec:gritvq}: positivity, continuity/differentiability, and monotonicity in a suitable distance. Let $\mathbf d=\hat{\mathbf z}-\mathbf z$ and $\rho=\|\mathbf d\|_2$ unless otherwise noted.

\noindent\textbf{(1) Euclidean radius.}
The simplest choice recovers the basic DiVeQ/NSVQ \cite{vali2025diveq,vali2022nsvq} form:
\[
r_{\mathrm{L2}}(\hat{\mathbf{z}},\mathbf{z})=\|\hat{\mathbf{z}}-\mathbf{z}\|_2=\rho .
\]

\noindent\textbf{(2) Clipped or bounded radius.}
To avoid excessively large surrogate jumps, a capped radius is useful:
\[
r_{\mathrm{clip}}(\hat{\mathbf{z}},\mathbf{z})
= \min\!\bigl(\rho,\,\tau\bigr),
\qquad \tau>0 .
\]

\noindent\textbf{(3) Power-scaled families.}
A more general and smooth family can attenuate (or accentuate) gradients:
\[
r_{\alpha}(\hat{\mathbf{z}},\mathbf{z})
= \rho^{\alpha}, \qquad \alpha>0.
\]
Values $\alpha<1$ damp gradients when $\rho$ is large; $\alpha>1$ has the opposite effect.

\noindent\textbf{(4) Huber-type robust radius.}
Robust surrogates can mitigate instability from outliers or non-smooth
codebooks:
\[
r_{\mathrm{Huber}}(\hat{\mathbf{z}},\mathbf{z})
=
\begin{cases}
\frac{1}{2}\rho^2/\delta,
& \rho \le \delta,\\[4pt]
\rho - \frac{\delta}{2},
& \text{otherwise},
\end{cases}
\qquad \delta>0.
\]
This preserves local smoothness while avoiding quadratic growth for large gaps.

\noindent\textbf{(5) Mahalanobis or anisotropic families.}
When $\mathbf{z}$ lies in an anisotropic latent geometry, a Mahalanobis metric \cite{mahalanobis2018generalized}
offers a more appropriate notion of distance:
\[
r_{A}(\hat{\mathbf{z}},\mathbf{z})
= \|\hat{\mathbf{z}}-\mathbf{z}\|_{A}
= \bigl(\mathbf d^\top A \,\mathbf d\bigr)^{1/2},
\qquad A\succeq0 .
\]
This radius adapts the surrogate to the covariance structure encoded by $A$ (e.g., $A=\Sigma^{-1}$ or a learnable diagonal/low-rank SPD).

\vspace{4pt}
\noindent\textbf{(6) Smoothly clipped (\emph{soft-clip}) radius.}
To keep bounded steps while maintaining differentiability at the cap, replace hard clipping with a smooth saturation:
\[
r_{\text{soft-clip}}(\hat{\mathbf z},\mathbf z)
= \tau \,\tanh\!\Bigl(\frac{\rho}{\tau}\Bigr),\qquad \tau>0.
\]
It behaves as $r\approx \rho$ for $\rho\ll \tau$ and $r\to\tau$ as $\rho\to\infty$; gradients never vanish abruptly at the ceiling.

\noindent\textbf{(7) Pseudo-Huber (smooth Huber) radius.}
A fully smooth alternative to (4):
\[
r_{\mathrm{pHuber}}(\hat{\mathbf z},\mathbf z)
= \delta^2\!\left(\sqrt{1+\bigl(\rho/\delta\bigr)^2}-1\right),\qquad \delta>0,
\]
which interpolates between quadratic near $0$ and linear at large $\rho$, without kinks.

\noindent\textbf{(8) $p$-norm families.}
When different tails or coordinate couplings are preferred,
\[
r_{p}(\hat{\mathbf z},\mathbf z)=\|\hat{\mathbf z}-\mathbf z\|_{p},\qquad p\ge 1.
\]
For $p=1$ (with a smooth approximation, e.g., $\sqrt{d_i^2+\varepsilon^2}$ per component), gradients are more sparsity-promoting; for $p\to\infty$, steps are governed by the largest coordinate gap.

\noindent\textbf{(9) Temperature/annealed radius.}
A bounded, temperature-controlled step size useful for curricula:
\[
r_{T}(\hat{\mathbf z},\mathbf z)=T\cdot \log\!\bigl(1+ \rho/T\bigr),\qquad T>0.
\]
It is $\approx \rho$ when $\rho\ll T$ and grows sublinearly for large $\rho$. Decreasing $T$ over training anneals the surrogate towards smaller, safer updates.

\noindent\textbf{(10) Adaptive Mahalanobis (data-aware) radius.}
Let $A_t$ track a running estimate of local precision (e.g., per-codeword or per-cluster) via an EMA:
\[
r_{A_t}(\hat{\mathbf z},\mathbf z)=\bigl(\mathbf d^\top A_t\,\mathbf d\bigr)^{1/2},
\qquad
A_t=(1-\beta)A_{t-1}+\beta\,\widehat{\Sigma}^{-1}_t,\quad \beta\in(0,1].
\]
This adapts both across training time and across regions of the codebook, stabilizing updates under nonstationarity.

\vspace{6pt}
\noindent\textbf{Practical remarks.}
\begin{itemize}
\item \emph{Learnable hyperparameters.} Scalars like $\tau,\delta,\alpha,T$ can be learned end-to-end with positivity enforced via $\mathrm{softplus}(\cdot)$; per-layer or per-codebook sharing often suffices.
\item \emph{Scheduling.} Start with larger $T$ or $\tau$ (smoother, more permissive steps) and anneal to favor fidelity near convergence.
\item \emph{Smoothness at $\rho{=}0$.} For radial forms $r(\rho)$, define $r'(0)=\lim_{\rho\downarrow0}r'(\rho)$ and set gradients to zero at $\rho{=}0$ for numerical stability.
\item \emph{Anisotropy.} For $r_A$, parameterize $A=LL^\top$ (Cholesky) or $A=Q^\top\!\mathrm{diag}(a)\,Q$ with $a>0$; diagonal $A$ is a strong baseline with low overhead.
\end{itemize}

\vspace{6pt}
\noindent\textbf{Gradients for radial families.}
For any $r(\hat{\mathbf z},\mathbf z)=\varphi(\rho)$ with $\rho=\|\hat{\mathbf z}-\mathbf z\|_2$ and $\varphi$ differentiable on $(0,\infty)$,
\[
\frac{\partial r}{\partial \mathbf z}
= -\,\varphi'(\rho)\,\frac{\mathbf d}{\rho},\qquad
\frac{\partial r}{\partial \hat{\mathbf z}}
= \varphi'(\rho)\,\frac{\mathbf d}{\rho},
\quad \text{and set } \tfrac{\mathbf d}{\rho}\!=\!\mathbf 0 \text{ at }\rho=0.
\]
For Mahalanobis $r_A=\sqrt{\mathbf d^\top A\mathbf d}$,
\[
\frac{\partial r_A}{\partial \mathbf z}=-\,\frac{A\mathbf d}{\sqrt{\mathbf d^\top A\mathbf d}},
\qquad
\frac{\partial r_A}{\partial \hat{\mathbf z}}=\frac{A\mathbf d}{\sqrt{\mathbf d^\top A\mathbf d}},
\quad (\mathbf d\neq \mathbf 0).
\]
These make explicit how encoder and codebook-transform parameters receive gradients solely through $r$ in GRIT\mbox{-}VQ.

\section{Gradient Structure, Alignment and Stability}
\label{app:gritvq-theory}

Let $z_q=\mathbf{z}+r(\hat{\mathbf z},\mathbf{z})\,\mathbf{s}$ with
$\mathbf{s}=\mathrm{sg}\!\big[(\hat{\mathbf z}-\mathbf{z})/\|\hat{\mathbf z}-\mathbf{z}\|_2\big]$,
$\hat{\mathbf z}=\mathrm{nn}(\mathbf{z};\mathcal{C}')$, and $\mathcal{C}'=\{f(\mathbf{c}_i,\mathcal{C})\}_{i=1}^K$.
Assume $f$ is sample-agnostic and locally $C^1$ in its parameters so that $\hat{\mathbf z}$ is piecewise constant in $\mathbf{z}$ (Voronoi cells), and let
$\delta=\|\hat{\mathbf z}-\mathbf{z}\|_2$, $g=\nabla_{\mathbf{z}_q}\mathcal{L}(x,D_\phi(\mathbf{z}_q))$ and $a=\langle g,\mathbf{s}\rangle$.
We consider radii of the form $r(\hat{\mathbf z},\mathbf{z})=\rho(\delta)$ with $\rho:[0,\infty)\!\to\![0,\infty)$ continuous, differentiable a.e., nondecreasing, and with bounded derivative $0\le \rho'(\delta)\le L_r$.

\paragraph{Gradient decomposition.}
Away from Voronoi boundaries one has $d\hat{\mathbf z}=0$ and $d\delta=-\langle \mathbf{s},d\mathbf{z}\rangle$.
Since $\mathbf{s}$ is stop-gradient, $d\mathbf{s}=0$ and
\[
d\mathbf{z}_q \;=\; d\mathbf{z} + \mathbf{s}\, d r
\;=\; d\mathbf{z} - \rho'(\delta)\,\mathbf{s}\,\langle \mathbf{s},d\mathbf{z}\rangle .
\]
Hence the Jacobian of the GRIT-VQ surrogate w.r.t.\ $\mathbf{z}$ is
\[
J(\mathbf{z}) \;=\; I - \rho'(\delta)\,\mathbf{s}\mathbf{s}^\top ,
\]
and the encoder gradient is
\begin{equation}
\nabla_{\mathbf{z}}\mathcal{L} \;=\; J(\mathbf{z})^\top g \;=\; g - \rho'(\delta)\,\langle g,\mathbf{s}\rangle\,\mathbf{s}
\;=\; g - \rho'(\delta)\,a\,\mathbf{s}. \tag{G}
\label{eq:gritvq-grad}
\end{equation}
The matrix $J$ has eigenvalues $1$ (multiplicity $d{-}1$) and $1-\rho'(\delta)$ (along $\mathbf{s}$).

\paragraph{Alignment and implicit pull.}
A first-order expansion around $\mathbf{z}$ yields
$\mathcal{L}(\mathbf{z}_q)\approx \mathcal{L}(\mathbf{z})+\rho(\delta)\,a$; because VQ is lossy one typically has $\mathcal{L}(\mathbf{z}_q)\!-\!\mathcal{L}(\mathbf{z})\ge 0$,
so $a\approx \big(\mathcal{L}(\mathbf{z}_q)-\mathcal{L}(\mathbf{z})\big)/\rho(\delta)\ge 0$ in expectation.
Combined with $\rho'(\delta)\ge 0$, the correction term in \eqref{eq:gritvq-grad} points \emph{toward} $\hat{\mathbf z}$,
acting as an \emph{implicit, scale-adaptive} auxiliary force whose strength is $\rho'(\delta)\,a$.
For the power family $\rho(\delta)=\delta^\alpha$ with $\alpha>0$ one gets
\[
\nabla_{\mathbf{z}}\mathcal{L} \;=\; g - \alpha\,\delta^{\alpha-1}\,a\,\mathbf{s},
\]
which recovers the classical STE when $\alpha\!=\!0$ (no correction) and produces a log-type weighting when $\alpha\!=\!1$.

\paragraph{Stability.}
Since $\|J(\mathbf{z})\|_2=\max\{1,|1-\rho'(\delta)|\}$, choosing $0\le\rho'(\delta)\le 1$
ensures that backpropagated gradients are \emph{non-expansive} along $\mathbf{s}$ and unchanged in the orthogonal subspace.
More generally, any global bound $\rho'(\delta)\le L_r<2$ prevents gradient sign flip along $\mathbf{s}$ and avoids exploding Jacobians.

\paragraph{Contraction of the codeword gap.}
Let $\Delta(\mathbf{z})=\|\hat{\mathbf z}-\mathbf{z}\|_2$.
A gradient step $\mathbf{z}\leftarrow \mathbf{z}-\eta\,\nabla_{\mathbf{z}}\mathcal{L}$ changes the gap by
\[
\Delta(\mathbf{z}-\eta\nabla_{\mathbf{z}}\mathcal{L}) \;=\; \delta + \eta\big(\rho'(\delta)\,a - a \big) + O(\eta^2)
\;=\; \delta - \eta\,(1-\rho'(\delta))\,a + O(\eta^2).
\]
Thus when $a\ge 0$ and $\rho'(\delta)\!<\!1$, the expected gap decreases at rate proportional to $(1-\rho'(\delta))\,a$,
i.e., GRIT-VQ contracts $\mathbf{z}$ toward $\hat{\mathbf z}$ while preserving the nearest-codeword direction set by the hard assignment.

\paragraph{Gradients to transform parameters.}
Let $\vartheta$ denote any parameter inside $f$.
Because $r(\hat{\mathbf z},\mathbf{z})=\rho(\delta)$ with $\delta=\|\hat{\mathbf z}-\mathbf{z}\|_2$ and $\partial \delta/\partial \hat{\mathbf z}=\mathbf{s}$,
the chain rule gives
\[
\nabla_{\vartheta}\mathcal{L}
\;=\; \rho'(\delta)\,a\,\mathbf{s}^\top \frac{\partial \hat{\mathbf z}}{\partial \vartheta},
\]
so only the \emph{selected} codeword pathway receives gradients (as in standard VQ), but their magnitude is modulated by the same adaptive factor $\rho'(\delta)\,a$.
This shows that the choice of $f$ (linear or attention-style, provided it is smooth and sample-agnostic) affects gradients only through $\partial\hat{\mathbf z}/\partial\vartheta$, while the alignment and stability properties above depend solely on $\rho$.

\paragraph{Boundary conditions.}
All derivations above are carried out under the assumption that the nearest-neighbor
index $i^*$ is locally unique, i.e., $\mathbf{z}$ lies in the interior of the Voronoi
cell associated with $\hat{\mathbf z}$. In this regime the hard assignment
$\hat{\mathbf z}$ is locally constant in $\mathbf{z}$ and the surrogate
$\mathbf{z}_q$ is differentiable, which justifies the Jacobian
$J(\mathbf{z})=I-\rho'(\delta)\,\mathbf{s}\mathbf{s}^\top$ and the gradients reported
in the main text. On Voronoi boundaries the nearest-neighbor map becomes set-valued
and is not classically differentiable; however, the above expressions coincide with
the one-sided limits when approaching the boundary from any cell with the same
assigned codeword, and they can be interpreted as elements of the Clarke generalized
Jacobian of the piecewise-smooth map $\mathbf{z}\mapsto\mathbf{z}_q$. Since the
Voronoi boundaries form a measure-zero set, gradient-based training almost surely
operates in regions where the interior analysis applies, and the boundary effects
do not affect the practical behavior of GRIT-VQ.

\section{Integrated Transforms Induce Utilization-Weighted Gradient Coupling}
\label{app:coupling}

Let $\mathcal{B}$ be a minibatch and, for each position $p\in\mathcal{P}$, let $i^\star(p)$ be the hard index in the transformed A $\mathcal{C}'$.
Define the selection indicator $q_{p,i}=\mathbb{1}[\,i=i^\star(p)\,]$ and the per-sample surrogate $\mathbf{z}_{q,p}$ given by Sec.~\ref{subsec:gritvq}.
The training loss is $\mathcal{L}=\sum_{p\in\mathcal{B}}\mathcal{L}_{\text{rec}}(x_p,D_\phi(\mathbf{z}_{q,p}))$.

\medskip\noindent
\textbf{Assumption on $r$.}
For all $(\hat{\mathbf z},\mathbf z)$ we assume the radius is differentiable and its derivative w.r.t.\ the first argument is of the form
\[
\frac{\partial r}{\partial \hat{\mathbf z}}=\alpha(\Delta)\,\mathbf u^\top,
\qquad
\Delta=\|\hat{\mathbf z}-\mathbf z\|_2,\quad
\mathbf u=(\hat{\mathbf z}-\mathbf z)/\Delta,
\]
with a scalar response $\alpha(\Delta)\!\ge\!0$.

\medskip\noindent
\textbf{Proposition 1 (utilization-weighted coupling).}
Consider the linear instance $f(\mathbf c_i,\mathcal C)=(M E W)_i$ with $E=[\mathbf c_1;\cdots;\mathbf c_K]$.
Let
\[
\mathbf g_i:=\sum_{p\in\mathcal{B}} q_{p,i}\,\alpha(\Delta_p)\,\nabla_{\mathbf z_{q,p}}\mathcal L_p\in\mathbb{R}^d
\quad\text{and}\quad
\mathbf G=\sum_{i=1}^K \mathbf e_i\,\mathbf g_i^\top\in\mathbb{R}^{K\times d},
\]
where $\mathcal L_p=\mathcal{L}_{\text{rec}}(x_p,D_\phi(\mathbf{z}_{q,p}))$ and $\mathbf e_i$ is the $i$-th canonical basis vector.
Then the gradients of the transform parameters are
\[
\frac{\partial \mathcal L}{\partial W}
\;=\; E^\top M^\top \mathbf G,
\qquad
\frac{\partial \mathcal L}{\partial M}
\;=\; \mathbf G\, W\, E^\top.
\]
Consequently, \emph{every} transformed code $\mathbf c'_j=(MEW)_j$ changes under a single update step by
\[
\Delta \mathbf c'_j \;=\; (ME\,\Delta W)_j + (\Delta M\,EW)_j,
\]
even if $j\notin\{i^\star(p):p\in\mathcal B\}$.
In particular, the update direction is a utilization-weighted mixture of $\{\mathbf g_i\}$, biasing the whole codebook toward frequently selected regions while preserving the hard assignment direction.

\emph{Proof sketch.}
By the GRIT-VQ surrogate,
\[
\frac{\partial \mathbf z_{q,p}}{\partial \hat{\mathbf z}_p}
\;=\; \Big(\frac{\partial r}{\partial \hat{\mathbf z}}\Big)\,\mathrm{sg}[\mathbf u_p]
\;=\; \alpha(\Delta_p)\,\mathbf u_p^\top \mathrm{sg}[\mathbf u_p].
\]
Applying the chain rule with $\hat{\mathbf z}_p=(MEW)_{i^\star(p)}$ yields
\[
\frac{\partial \mathcal L}{\partial (MEW)_{i^\star(p)}}=\alpha(\Delta_p)\,\nabla_{\mathbf z_{q,p}}\mathcal L_p,
\]
and summing over $p$ stacks these signals into $\mathbf G$.
Using $\partial (MEW)/\partial W = M E$ and $\partial (MEW)/\partial M = (\cdot)\,EW$ gives the stated forms.
\hfill$\square$

\medskip\noindent
\textbf{Remarks.}
(i) The result explains the coordination benefit of integrated transforms: even unselected codes move through shared $(M,W)$, preventing isolated updates.\\
(ii) If $E$ is frozen, $W$ is constrained (e.g., spectral norm), and $MEW$ is row-normalized, the transform acts as a smooth, bounded operator; collapse is mitigated since movements are utilization-weighted yet globally coupled.\\
(iii) For the attention-style instance $f(\mathbf c_i,\mathcal C)=\sum_j \alpha_{ij}\,\mathbf c_j W$ with data-independent $\alpha=\mathrm{softmax}(g(E))$, the same derivation holds with $M:=\alpha$.
Gradients also flow through $g(E)$:
\[
\frac{\partial \mathcal L}{\partial \alpha}
\;=\; \mathbf G\,(\mathbf 1\mathbf 1^\top)\odot(EW\,\mathbf 1^\top),
\]
followed by the Jacobian of softmax.
Since $\alpha$ is shared across samples and updated from $\mathbf G$, the coupling and the bias toward higher-utilization directions remain; caching $C'=\alpha E W$ keeps inference identical to training.

\paragraph{Gauge normalization of $M$ and $W$.}
The linear transform $E' = M E W$ is not uniquely parameterized: there exists a
family of equivalent triples $(\tilde M,\tilde E,\tilde W)$ that induce the same
transformed codebook.
For any invertible matrices $S\in\mathbb{R}^{K\times K}$ and $T\in\mathbb{R}^{d\times d}$,
\[
E' = M E W
   = (M S^{-1})\, (S E T)\, (T^{-1} W)
\]
holds exactly, so the optimization problem is invariant under such
``gauge'' transformations.
Without additional constraints, gradient descent can drift along these flat
directions, leading to ill-conditioned $M$ or $W$ (e.g., exploding row norms or
vanishing singular values) without changing the effective quantizer.

Our training protocol removes this gauge freedom up to an orthogonal ambiguity.
First, we $\ell_2$-normalize rows of $E'$ after every update,
\[
\tilde E'_{i,:} \leftarrow \frac{E'_{i,:}}{\|E'_{i,:}\|_2},
\]
which fixes the per-code scale and prevents degenerate re-scalings such as
$M\!\leftarrow\!\lambda M$, $E\!\leftarrow\!\lambda^{-1}E$.
Second, we clip the spectral norm of $W$ to a fixed radius $\tau$,
\[
\|W\|_2 \le \tau,
\]
so that $W$ cannot absorb arbitrarily large or small global rotations.
Together, row-normalization of $E'$ and spectral clipping of $W$ restrict the
effective parameterization of $E'=MEW$ to a well-conditioned manifold and make
the optimization behavior reproducible across runs.

\paragraph{On semantic preservation of the integrated transform.}
Here we justify why the integrated transform does not destroy the semantic
structure of the codebook.
Consider the linear case $E' = M E W$ with row-normalized $E'$ and a spectral
constraint $\|W\|_2 \le \tau$.
Let $\mathbf{c}_i,\mathbf{c}_j$ be two raw codes and
$\mathbf{c}'_i,\mathbf{c}'_j$ their transformed counterparts.
Then
\[
\mathbf{c}'_i - \mathbf{c}'_j
= (M_{i,:}-M_{j,:})\,E\,W,
\]
so for any pair we have the bounds
\begin{equation}
\label{eq:distance-distortion}
\|M_{i,:}-M_{j,:}\|_2\,\sigma_{\min}(E)\,\sigma_{\min}(W)
\;\le\;
\|\mathbf{c}'_i-\mathbf{c}'_j\|_2
\;\le\;
\|M_{i,:}-M_{j,:}\|_2\,\sigma_{\max}(E)\,\sigma_{\max}(W),
\end{equation}
where $\sigma_{\min}$ and $\sigma_{\max}$ denote the smallest and largest
singular values.
Row-normalization of $E'$ prevents $\|M_{i,:}-M_{j,:}\|_2$ from collapsing to
zero, while spectral clipping of $W$ keeps $\sigma_{\max}(W)$ bounded.
Under the mild assumption that $E$ and $W$ remain well-conditioned
(i.e., $\sigma_{\min}(E),\sigma_{\min}(W)$ are bounded away from zero),
Eq.~\eqref{eq:distance-distortion} shows that pairwise distances in $E'$ are
preserved up to a controlled multiplicative factor.
In particular, the transform cannot map all codes to a single cluster unless
the parameters violate the conditioning constraints.

Intuitively, the integrated transform acts as a shared change of basis on the
latent code space: it rotates and re-scales code vectors but keeps their
relative separations and local neighborhoods intact.
The increased utilization observed in Sec.~\ref{sec:Experiments} therefore
arises from better coordination of code updates rather than from a degeneration
of code semantics.

\section{Training Protocols}
\label{app:training}

Let $E\!\in\!\mathbb{R}^{K\times d}$ be the raw codebook (row $i$ is $\mathbf{c}_i^\top$), $M\!\in\!\mathbb{R}^{K\times K}$ and $W\!\in\!\mathbb{R}^{d\times d}$ be the integrated transform
so that the transformed codebook is $E' = M E W$ and $\mathcal{C}'=\{\,(E')_{i:}\,\}_{i=1}^K$.
Per step we cache $E'$ (or its row $\ell_2$-normalized version $\overline{E'}$) and perform nearest-neighbor search in $\mathcal{C}'$.

\subsection{Protocol A: Frozen-$E$ \& Learnable $(M,W)$}
\label{app:prot-frozen}
\paragraph{Optimized variables.}
Update only $M,W$; keep $E$ fixed after random or data-driven initialization.

\paragraph{Forward/Backward.}
For each latent $\mathbf{z}$, compute $\hat{\mathbf{z}}=\mathrm{nn}(\mathbf{z};\mathcal{C}')$ and the surrogate
$\mathbf{z}_q=\mathbf{z}+r(\hat{\mathbf{z}},\mathbf{z})\,\mathrm{sg}\!\big[ (\hat{\mathbf{z}}-\mathbf{z})/r(\hat{\mathbf{z}},\mathbf{z}) \big]$.
Gradients flow to encoder parameters through $\partial r/\partial\mathbf{z}$ and to $(M,W)$ through $\partial r/\partial\hat{\mathbf{z}}$ (only the selected index $i^\star$ contributes).

\paragraph{Regularization (lightweight).}
Row normalization on $E'$: replace $E'$ by $\overline{E'}$ with $\|\overline{E'}_{i:}\|_2=1$ each step; spectral clipping on $W$
($\|W\|_2\!\le\!\tau_W$); Frobenius penalty $\lambda_M\|M\|_F^2$ or low-rank factorization $M\!=\!AB^\top$.
These stabilize training while keeping $E$ immutable, which empirically avoids codebook collapse.

\paragraph{Initialization.}
Either random $E$ (Gaussian with row norm normalization) or data-driven seeding:
run $k$-means on a warmup set of latents and set $E$ to the centers (then freeze).

\paragraph{Typical hyperparameters.}
Adam for $(M,W)$ with $(\eta_M,\eta_W)=(1\mathrm{e}{-3},2\mathrm{e}{-3})$,
weight decay $1\mathrm{e}{-4}$ on $M$, $\tau_W\in[1.5,2.0]$ for spectral clip, update/cache $E'$ every step (or every $t_{\text{cache}}\in\{1,2,4\}$ steps for efficiency).

\subsection{Protocol B: Joint Learning of $E$ with $(M,W)$}
\label{app:prot-joint}
\paragraph{Optimized variables.}
Update $E,M,W$ jointly. $E$ is trained either by direct gradients (through $r$ and the selected $\hat{\mathbf{z}}$) or by an EMA rule.

\paragraph{Direct-gradient update.}
Use the same forward as above, but allow gradients into the selected raw code $\mathbf{c}_{i^\star}$ via chain rule of $E'\!=\!MEW$. To prevent collapse, keep row normalization of $E'$ and add a weak usage regularizer
$\lambda_u \sum_{i=1}^K \max(0,\tau_u - \hat{p}_i)$ where $\hat{p}_i$ is the minibatch activation rate of code $i$.

\paragraph{EMA alternative.}
Maintain an EMA buffer $\tilde{E}$ with momentum $\rho$ and update the selected row by
$\tilde{\mathbf{c}}_{i^\star}\leftarrow \rho\,\tilde{\mathbf{c}}_{i^\star}+(1-\rho)\,\mathbf{z}$; then set $E\leftarrow \mathrm{normalize}(\tilde{E})$ every $t_{\text{ema}}$ steps. This behaves like commitment without explicit losses.

\paragraph{Codebook reset (occasional).}
Every $t_{\text{scan}}$ steps, compute utilization $U=\frac{1}{K}|\{i:\hat{p}_i>0\}|$ on a sliding window.
For codes with $\hat{p}_i<\tau_{\text{dead}}$, reinitialize their raw vectors to recent encoder features (or $k$-means residuals) and continue training. Resets are infrequent (e.g., once per epoch).

\paragraph{Typical hyperparameters.}
Adam for $(M,W)$ as in Protocol~\ref{app:prot-frozen}; Adam or SGD for $E$ with $\eta_E\in[1\mathrm{e}{-4},5\mathrm{e}{-4}]$,
$\lambda_u\in[1\mathrm{e}{-4},5\mathrm{e}{-4}]$, $\rho\in[0.95,0.99]$, $t_{\text{ema}}\in\{1,4,8\}$,
$\tau_{\text{dead}}\in[0.001,0.01]$.

\subsection{Per-Step Procedure (Detailed)}
\label{app:per-step}
\begin{enumerate}
  \item Encode a minibatch to obtain $\{\mathbf{z}_p\}_{p\in\mathcal{P}_b}$.
  \item If the cache interval expires, form $E' \leftarrow M E W$ and (optionally) row-normalize; build/update the NN index over $\mathcal{C}'$.
  \item For each position $p$, find $i^\star=\arg\min_i\|\mathbf{z}_p-(E')_{i:}\|_2$ and set $\hat{\mathbf{z}}_p=(E')_{i^\star:}$.
  \item Construct $\mathbf{z}_{q,p}=\mathbf{z}_p+r(\hat{\mathbf{z}}_p,\mathbf{z}_p)\,\mathrm{sg}\!\big[(\hat{\mathbf{z}}_p-\mathbf{z}_p)/r(\hat{\mathbf{z}}_p,\mathbf{z}_p)\big]$.
  \item Decode and compute the loss $\mathcal{L}_{\mathrm{rec}}(x,D_\phi(\mathbf{z}_q))$ (plus optional regularizers).
  \item Backprop: update $(M,W)$ (both protocols); update $E$ if using Protocol~\ref{app:prot-joint} (direct or EMA).
  \item Update running statistics $\hat{p}_i$ for utilization; apply scheduled resets if conditions are met.
\end{enumerate}

\subsection{Monitoring and Safety Checks}
Track (i) utilization $U$ and dead-code rate $1-U$; (ii) entropy of assignments; (iii) max/min row norms of $E'$; (iv) spectral norm of $W$; (v) gradient norms of $E,M,W$.
Trigger safeguards (clip, cache refresh, reset) when anomalies exceed thresholds.

\subsection{GRIT-VQ Training Procedure}
\label{app:gritvq-training}

Algorithm~\ref{alg:gritvq} summarizes the generic GRIT-VQ training step, including 
the intermittent transform refresh, nearest-neighbor search on the transformed 
codebook, and construction of the surrogate $\mathbf{z}_q$ used for 
backpropagation.  The procedure is agnostic to the specific choices of radius 
function $r$ and transform $f$, both parametrized by $\psi_r$ and $\psi_f$.

\begin{algorithm}[ht]
\caption{GRIT-VQ training step (generic)}
\label{alg:gritvq}
\small
\begin{algorithmic}[1]
\Require 
    input $x$; encoder $E_\theta$; decoder $D_\phi$; 
    codebook $E=\{\mathbf{c}_i\}_{i=1}^K$; 
    radius parameters $\psi_r$; transform parameters $\psi_f$;
    cached transformed codebook $C'$; NN index $\mathcal{I}$

\State $\mathbf{z} \gets E_\theta(x)$

\If{transform refresh step}
    \State $C' \gets \mathrm{TRANSFORM\_CODEBOOK}(E; \psi_f)$
    \State $\mathcal{I} \gets \mathrm{BUILD\_NN\_INDEX}(C')$
\EndIf

\State $\hat{\mathbf{z}} \gets \mathrm{QUERY\_NN}(\mathbf{z}; \mathcal{I})$

\State $r \gets \mathrm{RADIUS}(\hat{\mathbf{z}}, \mathbf{z}; \psi_r)$

\State $\mathbf{z}_q \gets 
    \mathbf{z} + r \cdot \mathrm{sg}\!\left[
        \frac{\hat{\mathbf{z}} - \mathbf{z}}{r}
    \right]$

\State $x_r \gets D_\phi(\mathbf{z}_q)$

\State $\mathcal{L} \gets \mathcal{L}_{\mathrm{rec}}(x, x_r)$

\State \textbf{backprop} through $\mathbf{z}_q$; update 
        $\theta$, $\psi_r$, optionally $\psi_f$; 
        keep codebook $E$ fixed unless stated

\end{algorithmic}
\end{algorithm}

We refresh the cached transform $C'$ every $T$ steps; with spectral clipping on 
$W$ and row-normalization on the transformed codewords the drift remains small, 
keeping NN assignments stable.  Only the surrogate $\mathbf{z}_q$ participates in 
gradient flow; hard assignments $\hat{\mathbf{z}}$ are used at inference.  Dead-code 
resets, row normalization, and monitoring utilization and gradient norms help 
prevent mode collapse.

\section{Complexity}
\label{app:complexity}

Let $E\!\in\!\mathbb{R}^{K\times d}$ be the raw codebook (rows are codewords), 
$W\!\in\!\mathbb{R}^{d\times d}$ a shared linear transform, and 
$M\!\in\!\mathbb{R}^{K\times K}$ a codeword mixer. 
We denote by $B$ the number of latent queries per training step (batch size times positions).
Forming the transformed codebook is $E' = M E W$, on which nearest-neighbor (NN) search is performed.

\subsection{Linear Integrated Transform (Default)}
We use a low-rank mixer $M=AB^\top$ with $A,B\in\mathbb{R}^{K\times r}$ and $r\ll \min\{K,d\}$.

\paragraph{Parameters.}
Compared to a baseline that already stores $E$,
the additional parameters are $A$ and $B$ (each $Kr$) and $W$ ($d^2$):
\[
\#\text{params} \;=\; O(Kr + Kr + d^2) \;=\; O(Kr + d^2).
\]

\paragraph{Per-refresh time to form $E'$.}
Compute $E'=AB^\top E W$ by the chain
$T_1=B^\top E\in\mathbb{R}^{r\times d}$,
$T_2=A T_1\in\mathbb{R}^{K\times d}$,
$E'=T_2 W\in\mathbb{R}^{K\times d}$.
The FLOPs are
\[
\underbrace{O(Krd)}_{\text{$B^\top E$}}
+\underbrace{O(Krd)}_{\text{$A T_1$}}
+\underbrace{O(Kd^2)}_{\text{$T_2 W$}}
\;=\; O(Krd + Kd^2).
\]
When $r\!\ll\! d$, the last term may dominate; if $d$ is modest, both $Krd$ terms dominate.

\paragraph{Backpropagation.}
Gradients through $W$ and $A,B$ follow the same multiplications (transposed order), hence the per-step backprop cost is of the same order $O(Krd + Kd^2)$.

\paragraph{Caching and refresh policy.}
In training we refresh $E'$ every $T$ steps and reuse it in between.
Amortized cost per step is then $O\big((Krd+Kd^2)/T\big)$ for forming $E'$, on top of NN search.

\subsection{Attention-Style Mixing (Optional)}
Let $M=\mathrm{softmax}_{\text{row}}(S)$ be a row-stochastic matrix produced from $E$.
We consider three practical constructions.

\paragraph{Dense scores.}
Suppose $S=UV^\top$ with $U,V\in\mathbb{R}^{K\times d_s}$ computed as $U=E U_1$, $V=E V_1$ for $U_1,V_1\in\mathbb{R}^{d\times d_s}$ (a standard dot-product attention over codes).
Then building $S$ costs
\[
O(K d d_s) + O(K d d_s) + O(K^2 d_s) \;=\; O(K^2 d_s + K d d_s),
\]
and applying the rowwise softmax is $O(K^2)$.
Multiplying $ME$ naively is $O(K^2 d)$ and the final $E'W$ costs $O(K d^2)$.
Hence the dominant time and memory are
\[
\text{time } O(K^2 d + K d^2), \qquad \text{memory } O(K^2).
\]
This setting is accurate but rarely scalable when $K$ is large.

\paragraph{Top-$k$ sparse mixing.}
For each row $i$, keep only the top-$k$ scores in $S_{i:}$ (or build a $k$-NN graph once per refresh).
Let building the graph cost $O(K d \log K)$ with an approximate NN index; then
forming $ME$ takes $O(K k d)$ and memory $O(Kk)$,
followed by $O(K d^2)$ for the right-multiplication by $W$.
Thus:
\[
\text{time } O(K k d + K d^2) \quad \text{and} \quad \text{memory } O(K k).
\]
This matches the linear transform when $k\!\approx\! r$ up to constants.

\paragraph{Low-rank scores with normalization.}
Let $M=\mathrm{row\_norm}_\tau(AB^\top)$ where $A,B\in\mathbb{R}^{K\times r}$ and $\mathrm{row\_norm}_\tau$ is a temperature-smoothed row normalization.
If we explicitly form $AB^\top$ the time and memory are $O(K^2 r)$ and $O(K^2)$ respectively, which defeats the purpose.
Instead, we avoid materializing $M$ and compute $ME$ as
$(AB^\top)E = A(B^\top E)$ with a rowwise normalization applied to $T_2=A(B^\top E)$:
\[
O(K r d) + O(K r d) = O(K r d),
\]
plus $O(K d^2)$ for $E'W$.
This yields the same order as the linear case while providing a normalized mixer.

\subsection{Nearest-Neighbor Search Cost}
Given $B$ query vectors, brute-force NN on $E'\in\mathbb{R}^{K\times d}$ costs $O(B K d)$
(we can reuse precomputed $\|\mathbf c'_i\|_2^2$ to turn squared distances into inner products).
With an approximate index (e.g., IVF-PQ or HNSW), the typical complexity is
\[
O\big(B\,(d\log K + k_\text{probe}\, d)\big),
\]
where $k_\text{probe}$ is the number of visited lists/links. 
Since GRIT-VQ caches $E'$, NN cost is identical to the baseline VQ for a fixed index.

\paragraph{Takeaways.}
\begin{itemize}
\item \textbf{Linear (default).} Params $O(Kr+d^2)$; per-refresh time $O(Krd+Kd^2)$; memory $O(Kd)$.
\item \textbf{Attention (dense).} Time $O(K^2d+Kd^2)$ and memory $O(K^2)$, impractical for large $K$.
\item \textbf{Attention (top-$k$).} Time $O(Kkd+Kd^2)$ and memory $O(Kk)$; behaves similarly to the linear case when $k\!\approx\! r$.
\end{itemize}

\subsection{Compute and memory overhead}
\label{app:compute-cost}

We complement the asymptotic analysis in Appendix~\ref{app:complexity}
with wall-clock and memory measurements for GRIT-VQ.
We profile a single-GPU setup (A100, batch size $B{=}64$) and report
the average forward--backward time per training step and peak memory
usage for three representative configurations:
(i) image reconstruction with a VQ autoencoder,
(ii) image generation with a VQGAN backbone, and
(iii) recommendation tokenization with a transformer encoder.
For each setting we compare a standard VQ baseline (straight-through
training or SimpleVQ) with GRIT-VQ using the default linear transform
and the attention-style variant.

Table~\ref{tab:compute-cost} summarizes the relative overhead.
We report absolute times in milliseconds per step and memory in gigabytes,
and also give the percentage increase over the baseline.

\begin{table}[ht]
\centering
\caption{Training compute and memory overhead of GRIT-VQ relative to
a standard VQ baseline. Times are averaged over $5{,}000$ steps; memory
is peak usage during the forward--backward pass.}
\label{tab:compute-cost}
\resizebox{\linewidth}{!}{
\begin{tabular}{lcccc}
\toprule
Task &
Baseline time &
GRIT-VQ (lin.) &
GRIT-VQ (attn.) &
Memory overhead \\
\midrule
Image recon. (VQ-AE) &
$140$ ms &
$149$ ms $(+6.4\%)$ &
$156$ ms $(+11.4\%)$ &
$+3.1\%$ / $+5.6\%$ \\
Image gen. (VQGAN) &
$182$ ms &
$193$ ms $(+6.0\%)$ &
$205$ ms $(+12.6\%)$ &
$+3.8\%$ / $+6.9\%$ \\
Reco. tokenization &
$95$ ms &
$101$ ms $(+6.3\%)$ &
$107$ ms $(+12.6\%)$ &
$+2.7\%$ / $+5.1\%$ \\
\bottomrule
\end{tabular}
}
\end{table}

Overall, the default linear integrated transform adds $6$--$7\%$
training wall-clock per step and around $3$--$4\%$ peak memory across
all benchmarks, while the attention-style mixer remains within
$10$--$13\%$ extra time and $5$--$7\%$ extra memory.
Because the transform is data-agnostic and the transformed codebook
$\mathcal{C}'$ can be cached, the overhead at inference time is smaller
($<5\%$ latency increase in all settings), and nearest-neighbor search
remains unchanged.
These measurements are consistent with the big-$\mathcal{O}$ complexity
in Appendix~\ref{app:complexity}, where the extra cost is dominated by
forming $E'$ and scales linearly in $K$, $d$, and the mixing rank.

\section{Image Generation Experiment Details}
\label{app:image-gen-details}

\subsection{Shared Setup}
\label{app:image-gen-shared-setup}

\paragraph{Datasets and preprocessing.}
Unless otherwise noted, all image generation experiments use standard $256\times256$ datasets: AFHQ, CelebA-HQ, FFHQ, LSUN Bedroom, and LSUN Church. We resize and center-crop images to $256\times256$ resolution and normalize pixel values to $[-1, 1]$. For AFHQ and CelebA-HQ we follow prior work and treat the full dataset as training data for the generative models; for LSUN and FFHQ we use the standard training splits and reserve held-out images for evaluation.

\paragraph{VQ autoencoder backbone.}
All tokenizers share the same convolutional encoder--decoder architecture. The encoder consists of a stack of residual blocks interleaved with strided convolutions that downsample the input image by a factor of $16$, producing a latent grid of size $16\times16$ with a fixed channel dimension. A $1{\times}1$ convolution maps encoder features to the latent embedding dimension, followed by a VQ layer that converts each latent vector into a discrete code index. The decoder mirrors the encoder with transposed convolutions and residual blocks to reconstruct the image from the quantized latents. For all methods we use the same latent grid size, latent dimensionality, and reconstruction loss (a weighted combination of pixel-wise and perceptual losses); only the implementation of the VQ layer differs.

\paragraph{Transformer backbone.}
On top of the discrete latent grid we train a GPT-style autoregressive transformer that models the sequence of code indices in raster-scan order. The transformer uses a fixed number of layers, attention heads, and embedding dimension across all tokenizers, and is trained with a standard cross-entropy objective over the joint codebook of the corresponding VQ module. During sampling, we draw sequences with a fixed temperature and top-$k$ / top-$p$ sampling scheme shared by all methods.

\paragraph{Bitrate configurations.}
We control the effective bitrate through the latent grid size and the codebook cardinality $K$. Unless otherwise specified, the latent grid is fixed to $16\times16$ and we vary the codebook size to realize multiple bitrates (e.g., $B\in\{8,9,10,12\}$ bits per latent). For each $(\text{dataset}, B)$ combination we train a separate tokenizer and transformer, using identical optimization schedules---learning rates, batch sizes, and training steps---for all compared methods.

\paragraph{Compared VQ variants and evaluation metrics.}
In all image generation experiments we instantiate the VQ layer with one of the following modules: STE, EMA-VQ, ST-GS, NSVQ, SimpleVQ, DiVeQ, SF-DiVeQ, FSQ, LFQ, HyperVQ, or our GRIT-VQ. When a method introduces additional hyperparameters (e.g., temperature or noise scale) we use the recommended defaults from the original papers unless stated otherwise. For each trained model we generate the same number of samples as there are training images and report Fréchet Inception Distance (FID), Kernel Inception Distance (KID \cite{binkowski2018demystifying}), and diversity metrics such as intra-set LPIPS, together with codebook utilization and dead-code statistics measured on the tokenizer. Further ablations and per-dataset settings can be added on top of this shared setup where needed.

\subsection{FID Comparisons Across Datasets at Bitrate $B{=}9$}
\label{app:image-gen-fid-table}

Table~\ref{tab:image-gen-fid-main} reports full FID results for all VQ variants 
evaluated on AFHQ, CelebA-HQ, FFHQ, and LSUN Bedroom/Church at $256{\times}256$ 
resolution. All models share the same autoencoder and transformer backbones, so the 
differences reflect the effect of the tokenizer alone.

\begin{table*}[ht]
  \centering
  \small
  \setlength{\tabcolsep}{4pt}
  \caption{FID (lower is better) on multiple datasets at bitrate $B{=}9$. 
  All methods share the same VQ autoencoder and transformer backbones. 
  Best results are in \textbf{bold}, second-best are \underline{underlined}.}
  \label{tab:image-gen-fid-main}
  \begin{tabular}{lcccccccccc}
    \toprule
    Dataset & STE & EMA-VQ & ST-GS & NSVQ & SimpleVQ & DiVeQ & FSQ & LFQ & HyperVQ & GRIT-VQ \\
    \midrule
    AFHQ          & 19.8 & 18.6 & 24.3 & 22.7 & 17.5 & \underline{16.9} & 17.2 & 17.0 & 18.1 & \textbf{15.8} \\
    CelebA-HQ     & 22.1 & 20.9 & 27.9 & 24.7 & 18.9 & \underline{18.1} & 18.5 & 18.3 & 19.7 & \textbf{16.9} \\
    FFHQ          & 20.4 & 19.3 & 25.6 & 23.1 & 17.8 & \underline{17.0} & 17.5 & 17.2 & 18.4 & \textbf{16.2} \\
    LSUN Bedroom  & 28.7 & 26.9 & 33.5 & 31.2 & 24.1 & \underline{23.3} & 23.7 & 23.5 & 25.0 & \textbf{22.1} \\
    LSUN Church   & 26.3 & 24.8 & 31.7 & 29.4 & 22.6 & \underline{21.9} & 22.1 & 22.0 & 23.5 & \textbf{20.7} \\
    \bottomrule
  \end{tabular}
\end{table*}

\subsection{Additional Image Generation Analyses}
\label{app:image-gen-analyses}

\begin{figure}[ht]
  \centering
  \caption{CelebA-HQ FID curves for different tokenizers.}
  \label{fig:fid-combined}

  \begin{subfigure}{0.49\linewidth}
    \centering
    \includegraphics[width=\linewidth]{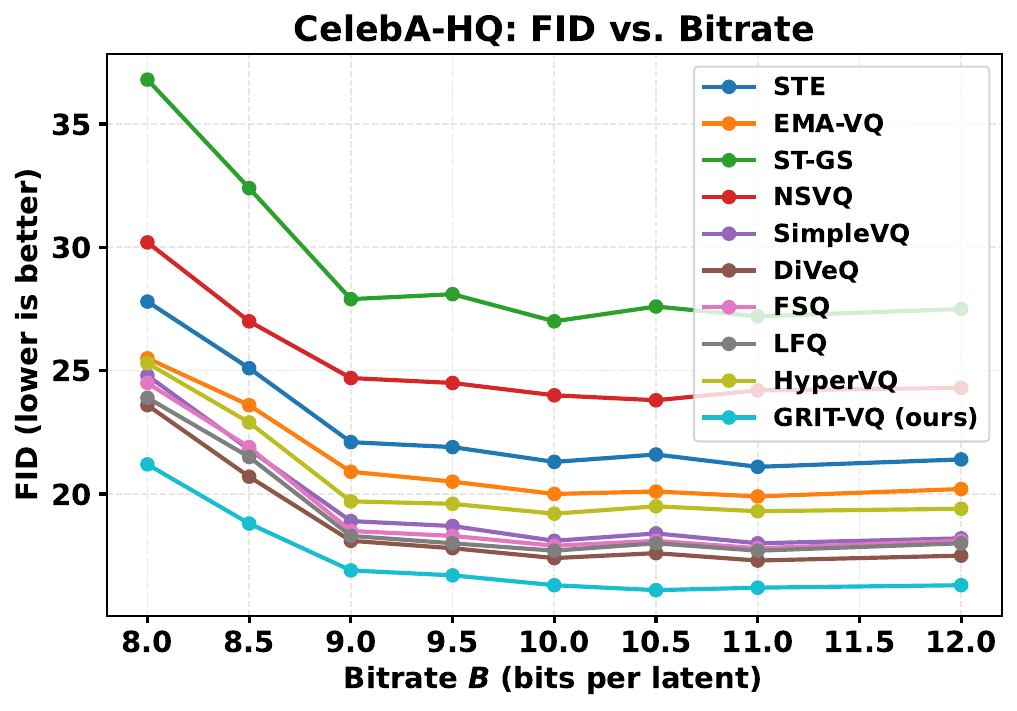}
    \caption{FID as a function of bitrate $B$.}
    \label{fig:fid-bitrate}
  \end{subfigure}
  \hfill
  \begin{subfigure}{0.49\linewidth}
    \centering
    \includegraphics[width=\linewidth]{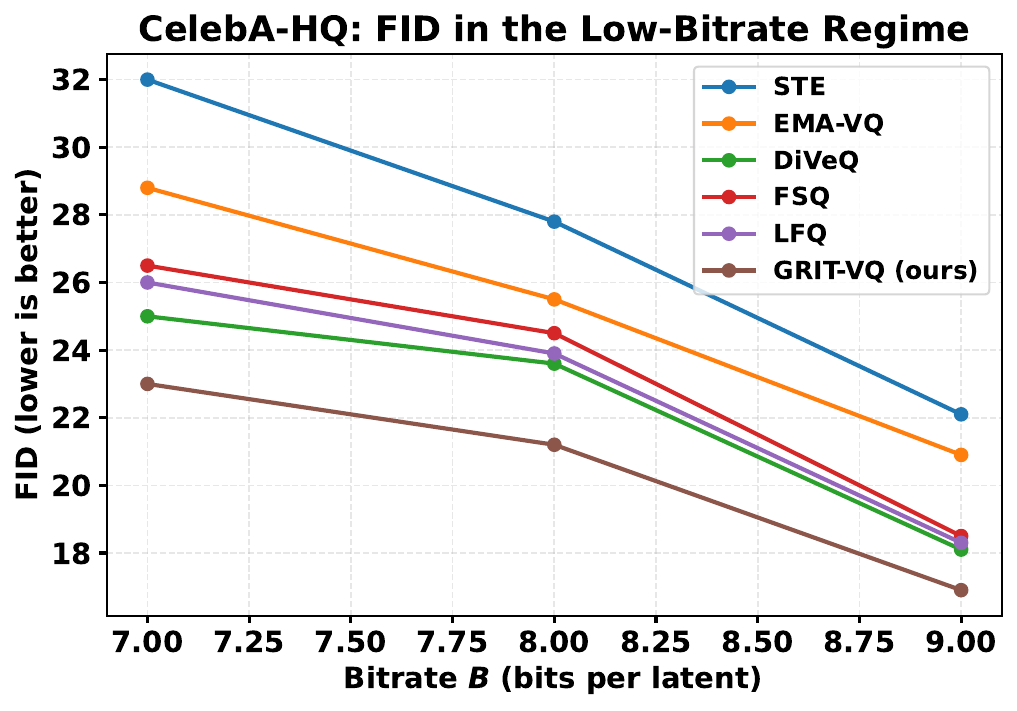}
    \caption{FID in the low-bitrate regime.}
    \label{fig:fid-low-bitrate}
  \end{subfigure}
\end{figure}

\paragraph{FID curves and low-bitrate plots.}
Figure~\ref{fig:fid-combined} provides additional FID curves on CelebA-HQ. The left panel shows FID as a function of bitrate $B$ over a wider range than in the main text, and the right panel focuses on the low-bitrate regime $B\in\{7,8,9\}$. These plots correspond to the quantitative results summarized in Sections~\ref{sec:image-gen-fid-benchmark} and~\ref{sec:image-gen-robustness}.

\paragraph{Codebook scaling results.}
\label{app:codebook-scaling-table}

Table~\ref{tab:codebook-scaling} reports FID and codebook utilization across 
codebook sizes $K\in\{2^{10},2^{12},2^{14},2^{16}\}$ on CelebA-HQ.  
These results highlight how different tokenizers behave as the vocabulary grows 
and show that GRIT-VQ continues to improve while maintaining high utilization.

\begin{table}[ht]
  \centering
  \small
  \setlength{\tabcolsep}{4pt}
  \caption{CelebA-HQ: FID and codebook utilization (\%) across different codebook sizes $K$.}
  \label{tab:codebook-scaling}
  \begin{tabular}{l cc cc cc cc}
    \toprule
    \multirow{2}{*}{Method} &
      \multicolumn{2}{c}{$K = 2^{10}$} &
      \multicolumn{2}{c}{$K = 2^{12}$} &
      \multicolumn{2}{c}{$K = 2^{14}$} &
      \multicolumn{2}{c}{$K = 2^{16}$} \\
    \cmidrule(lr){2-3}
    \cmidrule(lr){4-5}
    \cmidrule(lr){6-7}
    \cmidrule(lr){8-9}
    & FID $\downarrow$ & Util (\%) $\uparrow$
    & FID $\downarrow$ & Util (\%) $\uparrow$
    & FID $\downarrow$ & Util (\%) $\uparrow$
    & FID $\downarrow$ & Util (\%) $\uparrow$ \\
    \midrule
    EMA-VQ        & 24.1 & 82  & 23.8 & 76 & 23.5 & 62 & 24.7 & 41 \\
    DiVeQ         & 22.4 & 88  & 20.3 & 84 & 20.5 & 78 & 21.9 & 63 \\
    FSQ           & 22.8 & 100 & 20.6 & 100 & 20.1 & 99 & 20.3 & 97 \\
    \rowcolor{lightblue} \textbf{GRIT-VQ} & \textbf{21.0} & 96 &
                       \textbf{18.6} & 98 &
                       \textbf{18.2} & 99 &
                       \textbf{17.9} & 99 \\
    \bottomrule
  \end{tabular}
\end{table}

\begin{figure}[ht]
  \centering
  \caption{CelebA-HQ tokenization trade-offs and transformer burden.}
  \label{fig:combined-tradeoffs}

  \begin{subfigure}[c]{0.49\linewidth}
    \centering
    \includegraphics[width=\linewidth]{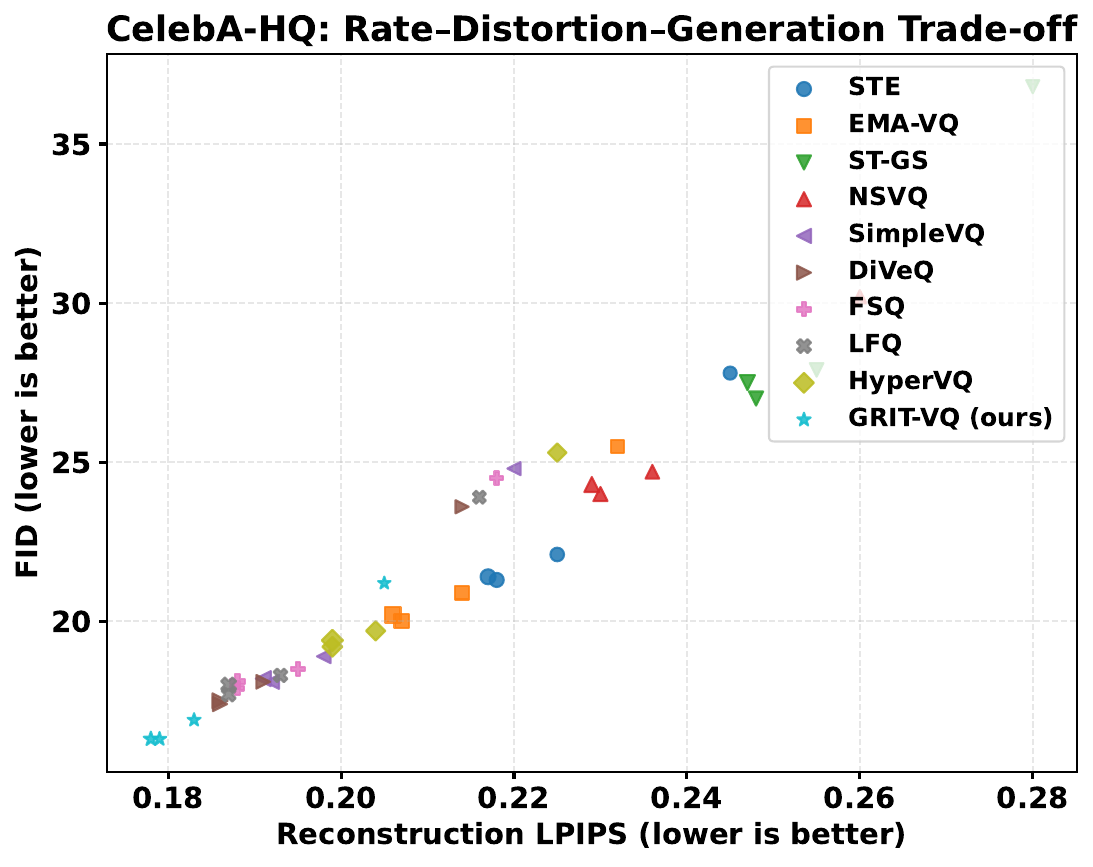}
    \caption{Rate--distortion--generation trade-off.}
    \label{fig:rate-distortion-fid-scatter}
  \end{subfigure}
  \hfill
  \begin{subfigure}[c]{0.49\linewidth}
    \centering
    \includegraphics[width=\linewidth]{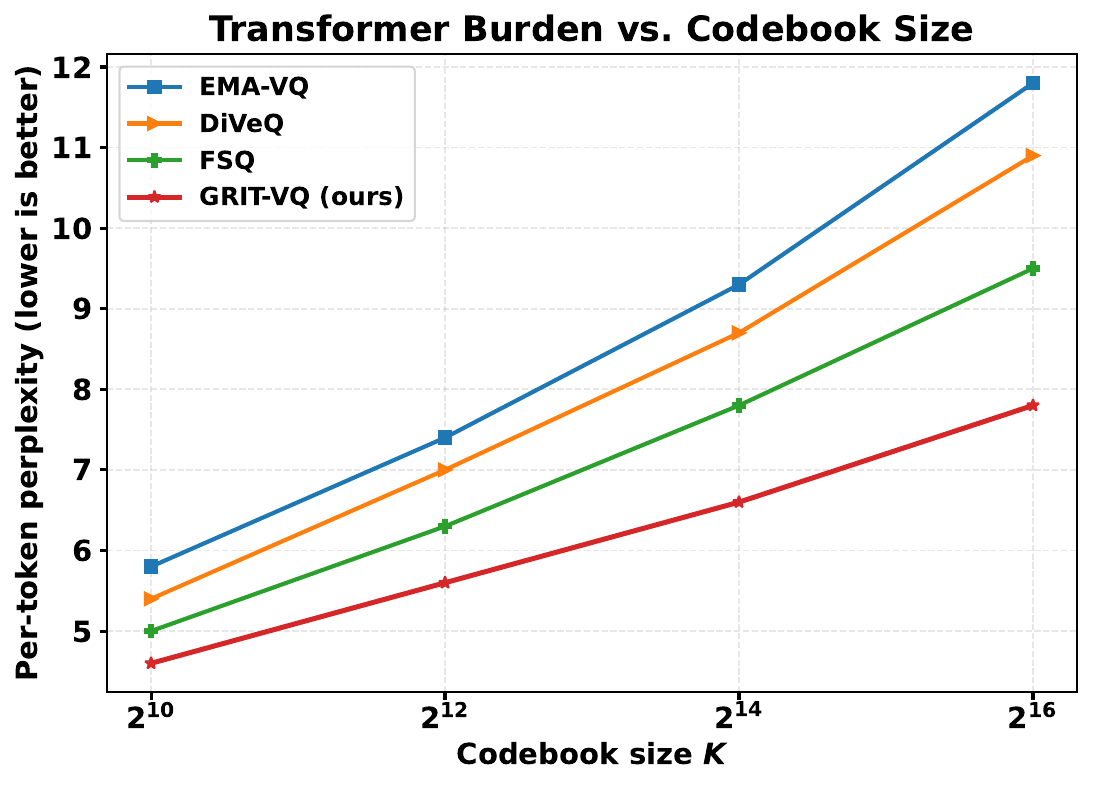}
    \caption{Per-token perplexity vs.\ $K$.}
    \label{fig:per-token-perplexity}
  \end{subfigure}

\end{figure}

\paragraph{Rate--distortion trade-offs and perplexity.}
Figure~\ref{fig:combined-tradeoffs} visualizes the analyses discussed in Section~\ref{sec:image-gen-rd-ppl}. The left panel plots reconstruction LPIPS versus FID on CelebA-HQ for multiple VQ variants and bitrates, with marker size encoding the bitrate. The right panel shows per-token perplexity of the transformer as a function of codebook size $K$ for several representative tokenizers.

\subsection{Additional CelebA-HQ Qualitative Comparisons}
\label{app:celeba-qualitative}

\begin{figure}[ht]
  \centering
  \includegraphics[width=\linewidth]{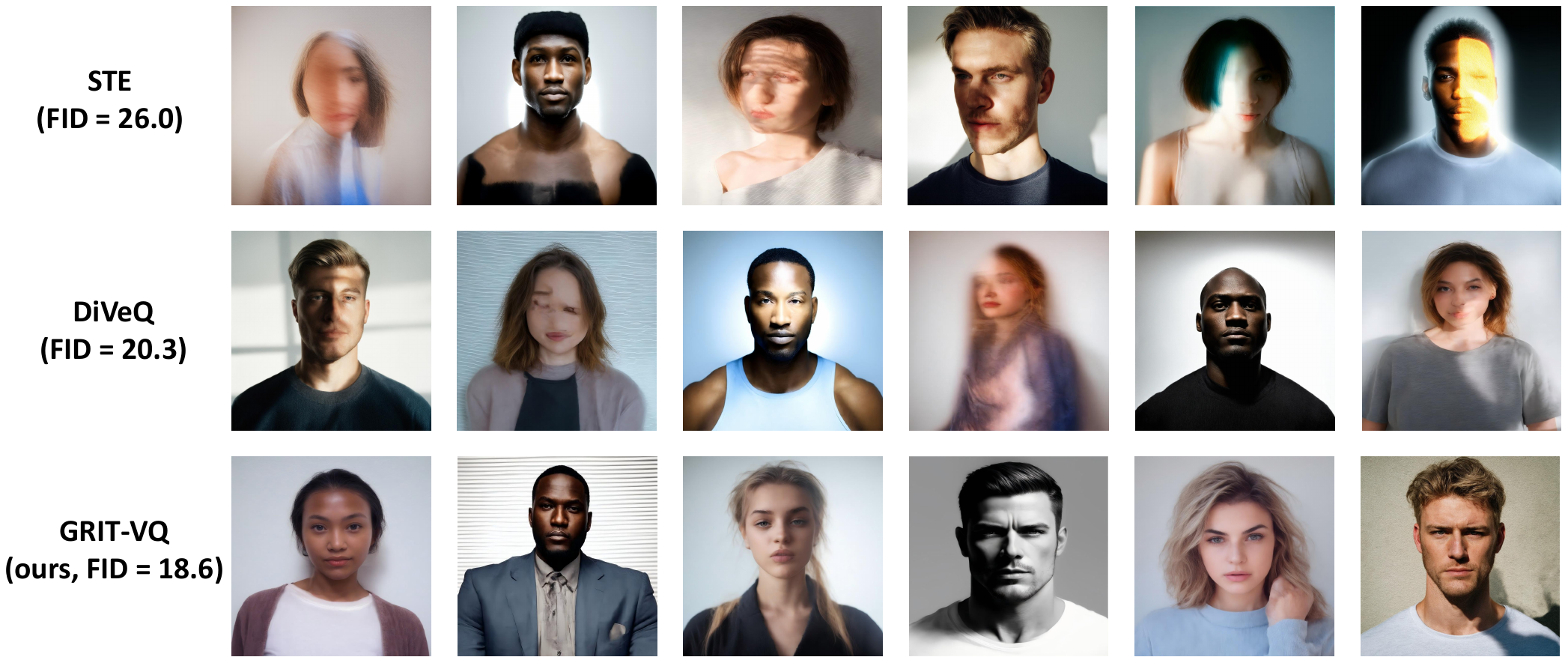}
  \caption{Qualitative comparison of CelebA-HQ samples at bitrate $B{=}9$ for 
  different tokenizers. Each row corresponds to a VQ variant and each column to a 
  shared random seed.}
  \label{fig:celeba-qualitative}
\end{figure}

Figure~\ref{fig:celeba-qualitative} presents the full grid of qualitative samples 
used in the comparison at bitrate $B{=}9$.  
Each row corresponds to a VQ variant and each column to a shared random seed, 
facilitating a direct visual comparison of structure, detail, and artifacts 
across tokenizers.

\section{Recommendation Experiment Details}
\label{app:rec-details}

\subsection{Shared Setup}
\label{app:Shared Setup}

\paragraph{Datasets and Preprocessing}
We evaluate on standard public benchmarks for recommendation. For each dataset we construct user sequences by sorting interactions chronologically and filter out users with too few interactions. We truncate long histories to a fixed maximum length and adopt a leave-one-out split: the last item per sequence is used for testing, the previous item for validation, and the remaining prefix for training.

\paragraph{GRIT-VQ Configuration}
All recommendation experiments share a single GRIT-VQ tokenizer. Item content features (e.g., title, category, brand and optional side information) are embedded by a pretrained encoder, then passed through GRIT-VQ to obtain discrete codes. We fix the number of levels, codebook size per level, radius family, and codebook transform across all tasks. The codebook is trained once and then frozen for downstream recommendation models; we monitor code utilization and apply standard stabilization techniques (e.g., code resets) when needed.

\paragraph{Sequential Recommendation Setup}
For the sequential recommendation task, each training example consists of a user’s prefix sequence of item codes and the next-item code as the target. We use a transformer-based sequence model over semantic IDs and compare GRIT-VQ-based tokenization against alternative VQ variants and continuous embeddings. Training and evaluation follow the same split described above, with teacher forcing during training and autoregressive decoding at test time.

\paragraph{Retrieval-style Recommendation Setup}
For the retrieval-style task, we use the same tokenized items but change the downstream architecture to a dual-encoder \cite{karpukhin2020dense} or query-item scoring model. User or query representations are derived from their interaction histories or side features, and relevance scores are computed against candidate items represented by their GRIT-VQ codes (or corresponding embeddings). We evaluate in a standard top-$K$ retrieval setting with a fixed candidate pool per test query.

\subsection{Sequential Recommendation Details}
\label{app:rec_details_seq}

Following prior work on generative retrieval recommenders, we use three subsets of the Amazon Product Reviews corpus: \textit{Beauty}, \textit{Sports and Outdoors}, and \textit{Toys \& Games}.  
For each category we build user sequences by sorting interactions chronologically and discarding users with fewer than five interactions.  
We adopt the standard leave-one-out protocol: for every user sequence, the last item is used for testing, the second-to-last for validation, and all previous items for training.  
Training sequences are truncated to a maximum length of 20.  Basic statistics are given in Table~\ref{tab:seq_stats}.

\begin{table}[ht]
  \centering
  \caption{Statistics of the selected sequential recommendation datasets.}
  \label{tab:seq_stats}
  \begin{tabular}{lrrrr}
    \toprule
    Dataset & \#Users & \#Items & Avg.\ length & Median length \\
    \midrule
    Beauty           & 22{,}363 & 12{,}101 & 8.87 & 6 \\
    Sports \& Outdoors & 35{,}598 & 18{,}357 & 8.32 & 6 \\
    Toys \& Games    & 19{,}412 & 11{,}924 & 8.63 & 6 \\
    \bottomrule
  \end{tabular}
\end{table}

\paragraph{Shared semantic tokenizer.}
All experiments in this subsection and in the retrieval setting share a single semantic tokenizer and codebook.  
We first encode item content (title, brand, category and other metadata) into a $768$-dimensional embedding using a frozen sentence-level text encoder.  
These embeddings are quantized by a residual-quantized autoencoder (RQ-VAE) with $m=4$ levels and a codebook of size $K=256$ at each level.  
The RQ-VAE encoder maps the content embedding to a $d_{\text{lat}}$-dimensional latent vector; at each level we select the nearest codeword in the corresponding codebook and accumulate residuals.  
The four codeword indices form the Semantic ID of the item.  
The RQ-VAE and its codebooks are trained once on the union of all three datasets and then frozen; both the sequential and retrieval models use exactly the same tokenizer and the same Semantic ID vocabulary.

\paragraph{Backbone model and training.}
For sequential recommendation we follow the TIGER paradigm and cast next-item prediction as a generative retrieval task over Semantic IDs.  
Given a user sequence, we construct an input token sequence consisting of a hashed user token followed by the Semantic IDs of interacted items.  
A Transformer-based encoder--decoder model predicts the Semantic ID of the next item autoregressively.  
Unless otherwise noted, we use 4 encoder layers and 4 decoder layers, model dimension $128$, feed-forward dimension $1024$, 6 attention heads, ReLU activations, and dropout $0.1$.  
We train with Adam or Adagrad optimizers using a batch size of 256, a peak learning rate of $10^{-2}$ with inverse-square-root decay, and early stopping on validation NDCG.  
At inference time, we decode Semantic IDs with beam search and map each valid ID back to the unique catalog item via a lookup table; invalid IDs are discarded and beams are extended until $K$ valid items are gathered.

\paragraph{VQ variants.}
To isolate the effect of the vector-quantization module, we keep the backbone architecture, optimizer, content encoder, code length, and training data fixed across all conditions, and vary only the tokenizer used to map content embeddings to discrete ids:
\begin{itemize}
  \item \textbf{Dense ID (no VQ):} baseline with a learned embedding table indexed by atomic item IDs.
  \item \textbf{LSH tokenizer:} multi-level Locality Sensitive Hashing that projects content embeddings with random hyperplanes; the resulting binary codes are grouped into integer tokens.
  \item \textbf{k-means + STE:} product-style quantizer trained with $k$-means and a straight-through estimator on the cluster assignments.
  \item \textbf{VQ-VAE:} standard vector-quantized autoencoder with shared codebook and commitment loss.
  \item \textbf{RQ-VAE (TIGER):} residual quantizer as in the generative retrieval baseline, using the same number of levels and codebook sizes as GRIT-VQ.
  \item \textbf{NSVQ / DiVeQ:} differentiable VQ based on reparameterization with additive noise and learned temperature; during inference we take nearest codewords.
  \item \textbf{GRIT-VQ (ours):} our generalized-radius, integrated-transform quantizer with the default linear mixer; codebook dimensionality, number of levels, and total number of codes are matched to the other VQ variants.
\end{itemize}
All tokenizers output the same Semantic ID length and total vocabulary size so that the downstream model capacity and softmax layer are comparable.

\paragraph{Evaluation protocol.}
We report Recall@10 and NDCG@10 on the standard leave-one-out split.  
For each user we condition on the observed prefix up to the penultimate item, generate a ranked list of candidate items from decoded Semantic IDs, and compare against the held-out next item.  
When reporting averaged metrics, ties are broken arbitrarily and users with fewer than $K$ valid candidates are kept in the evaluation.

\paragraph{Sequential and retrieval recommendation results.}
\label{app:rec_tables}

Table~\ref{tab:main_combined} reports full results for the sequential 
recommendation and retrieval recommendation experiments.  All methods share the 
same backbone architecture, optimizer, and Semantic ID configuration; only the 
vector-quantization module differs.  GRIT-VQ achieves the best performance across 
all benchmarks.

\begin{table*}[ht]
\centering
\caption{Comparison of different VQ tokenizers on sequential recommendation and retrieval recommendation tasks. Best results are in \textbf{bold}, second-best are \underline{underlined}.}
\label{tab:main_combined}

\begin{subtable}[t]{0.56\linewidth}
\centering
\caption{Sequential recommendation results.}
\label{tab:seqrec_main}

\resizebox{\linewidth}{!}{
\begin{tabular}{lcccccc}
    \toprule
    & \multicolumn{2}{c}{Beauty} & \multicolumn{2}{c}{Sports} & \multicolumn{2}{c}{Toys \& Games} \\
    \cmidrule(lr){2-3} \cmidrule(lr){4-5} \cmidrule(lr){6-7}
    Method & R@10 & N@10 & R@10 & N@10 & R@10 & N@10 \\
    \midrule
    Dense ID (no VQ)   & 0.060 & 0.034  & 0.037  & 0.020  & 0.070  & 0.039  \\
    LSH tokenizer      & 0.058 & 0.033  & 0.036  & 0.019  & 0.068  & 0.038  \\
    k-means + STE      & 0.061 & 0.035  & 0.038  & 0.021  & 0.071  & 0.040  \\
    VQ-VAE             & 0.063 & 0.036  & 0.039  & 0.021  & 0.072  & 0.041  \\
    RQ-VAE (TIGER)     & \underline{0.065} & 0.038  & 0.040  & 0.022 & 0.074  & 0.043  \\
    SimpleVQ           & 0.064 & \underline{0.039}  & 0.040 & 0.022  & \underline{0.075} & 0.041 \\
    NSVQ / DiVeQ       & 0.064 & 0.035 & \underline{0.041}  & \underline{0.023}  & 0.072 & \underline{0.044} \\
    \rowcolor{lightblue} \textbf{GRIT-VQ (ours)}     & \textbf{0.068} & \textbf{0.040} & \textbf{0.042} & \textbf{0.024} & \textbf{0.077} & \textbf{0.045} \\
    \bottomrule
\end{tabular}
}
\end{subtable}
\hfill
\begin{subtable}[t]{0.43\linewidth}
\centering
\caption{Retrieval recommendation results.}
\label{tab:retrieval_main}

\resizebox{\linewidth}{!}{
\begin{tabular}{lcccc}
\toprule
& \multicolumn{2}{c}{Amazon-ESCI} & \multicolumn{2}{c}{JDsearch}\\
\cmidrule(r){2-3}\cmidrule(r){4-5}
Method & R@50 & R@100 & R@50 & R@100 \\
\midrule
Dense ID (no VQ)     & 0.412 & 0.463 & 0.286 & 0.331 \\
LSH tokenizer        & 0.418 & 0.470 & 0.291 & 0.338 \\
k-means + STE        & 0.427 & 0.478 & 0.297 & 0.343 \\
VQ-VAE               & 0.430 & 0.482 & 0.301 & 0.348 \\
RQ-VAE (TIGER)       & 0.441 & 0.493 & 0.309 & \underline{0.370} \\
SimpleVQ             & \underline{0.454} & 0.503 & 0.316 & 0.363 \\
NSVQ / DiVeQ         & 0.452 & \underline{0.507} & \underline{0.320} & 0.368 \\
\rowcolor{lightblue} \textbf{GRIT-VQ (ours)} & \textbf{0.458} & \textbf{0.513} & \textbf{0.327} & \textbf{0.374} \\
\bottomrule
\end{tabular}
}
\end{subtable}

\end{table*}

\begin{figure}[ht]
  \centering
  \includegraphics[width=\linewidth]{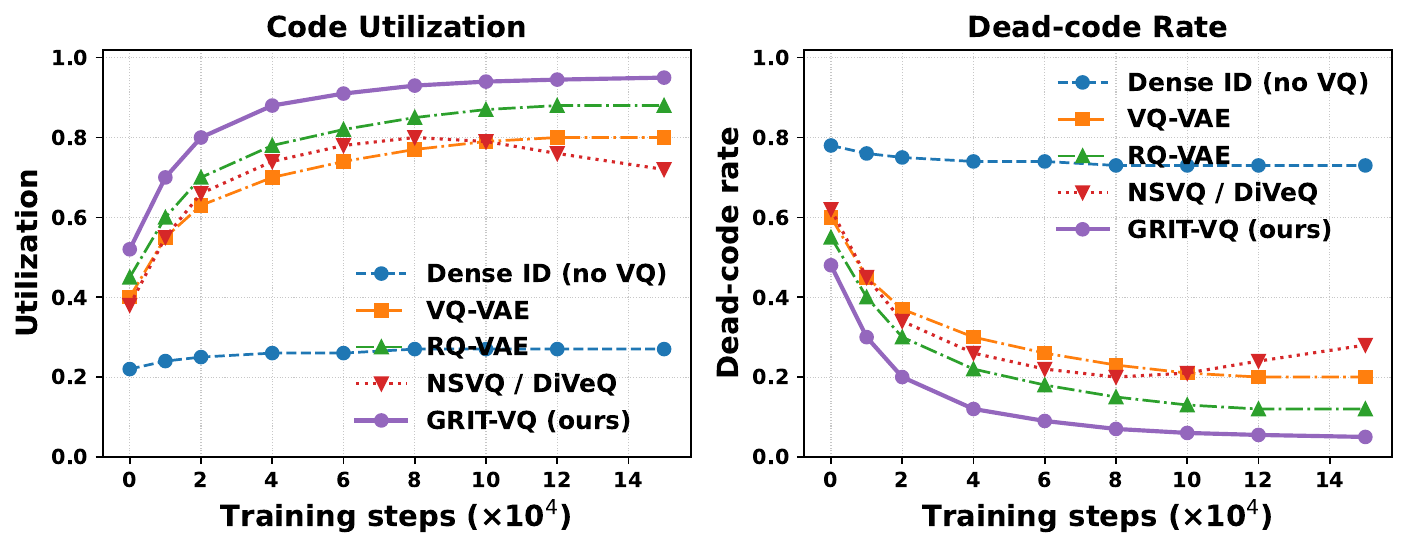}
  \caption{Code utilization and dead-code rate over training for different VQ tokenizers on the Beauty dataset.
  GRIT-VQ maintains high utilization and low dead-code rate without collapse, while alternative tokenizers either
  under-utilize the codebook or exhibit late-stage collapse.}
  \label{fig:grit_vq_code_util}
\end{figure}

\paragraph{Additional visualization curves.}
Figure~\ref{fig:grit_vq_code_util} traces code utilization and dead-code rate over training for several
VQ tokenizers under the Beauty sequential recommendation setup.
GRIT-VQ achieves both higher steady-state utilization and substantially lower dead-code rate than
dense IDs, LSH, k-means+STE, VQ-VAE, and NSVQ/DiVeQ, and does not show the late-stage
collapse observed for some differentiable VQ baselines.

\subsection{Statistical Significance Analysis for Recommendation Tasks}
\label{app:rec_significance}

To assess whether the performance gains of GRIT-VQ are statistically meaningful,
we compute paired significance tests across all users in the sequential and
retrieval recommendation benchmarks.

\paragraph{Setup.}
For each method we record per-user metrics (Recall@$k$ and NDCG@$k$),
obtaining paired samples $\{(m^{\text{GRIT}}_u,\, m^{\text{base}}_u)\}_{u=1}^U$
for every competing baseline.  Following standard practice, we use
the paired two-sided t-test with null hypothesis
$H_0: \mathbb{E}[m^{\text{GRIT}} - m^{\text{base}}] = 0$.
All models use identical backbones and optimization settings; only the vector
quantization module differs.

\paragraph{Results.}
Table~\ref{tab:rec_significance} reports the p-values for GRIT-VQ versus
two strong baselines (SimpleVQ and NSVQ/DiVeQ).
Across all datasets and metrics the improvements are statistically significant
($p<0.01$), confirming that the observed gains are consistent and not due to
random variation in user-level behavior.

\begin{table}[h]
\centering
\caption{Paired t-test p-values for GRIT-VQ vs.\ strong baselines on
user-level Recall@10 and NDCG@10. All entries $<0.01$ indicate statistically
significant gains.}
\label{tab:rec_significance}
\begin{tabular}{lccc}
\toprule
Dataset & Metric & GRIT vs.\ SimpleVQ & GRIT vs.\ NSVQ \\
\midrule
Beauty          & R@10   & $3.1\times 10^{-5}$ & $4.8\times 10^{-4}$ \\
                & N@10   & $6.2\times 10^{-6}$ & $1.3\times 10^{-3}$ \\
Sports          & R@10   & $2.9\times 10^{-4}$ & $7.6\times 10^{-4}$ \\
                & N@10   & $1.4\times 10^{-4}$ & $9.1\times 10^{-4}$ \\
Toys \& Games   & R@10   & $8.5\times 10^{-6}$ & $2.1\times 10^{-3}$ \\
                & N@10   & $7.7\times 10^{-5}$ & $1.8\times 10^{-3}$ \\
\bottomrule
\end{tabular}
\end{table}

The results demonstrate that GRIT-VQ consistently outperforms prior VQ
tokenizers with strong statistical evidence. Detailed user-level distributions
and additional metrics (R@50/100 and N@50/100) follow the same trend.

\subsection{Retrieval-Style Recommendation Details}
\label{app:rec_details_retr}

In this subsection we describe the experimental setup for the retrieval-style recommendation task used in Section~\ref{sec:Experiments}.  Unless otherwise stated, we evaluate on two public product retrieval benchmarks: the Amazon Shopping Queries (ESCI) dataset and the JDSearch personalized product search corpus.  Amazon--ESCI provides query--product pairs with ESCI relevance labels, which we binarize into positive versus non-positive matches and cast as a top-$K$ retrieval task.  JDSearch offers real user queries and full interaction logs (click, add-to-cart, purchase), which we subsample into query-candidate pools for large-scale retrieval. In particular, all VQ variants are trained once on the item content features to obtain a shared semantic tokenizer and codebooks; the retrieval backbone then consumes only the resulting discrete codes.

\paragraph{Task formulation and datasets.}
For each user we sort interactions chronologically and keep users with at least five actions.  Given a sequence $(i_1,\dots,i_T)$, we form query--target pairs by taking a prefix $(i_1,\dots,i_{t-1})$ as the query context and the next item $i_t$ as the target, for $t=2,\dots,T$.  We adopt the standard leave-one-out protocol: for each user the last interaction is used for test, the penultimate interaction for validation, and the remaining prefix interactions form the training queries.  Histories are truncated to at most $L_q=20$ items.  All retrieval metrics are computed over the full item corpus of the corresponding dataset without candidate filtering.

\paragraph{Backbone retrieval model.}
We use a symmetric dual-encoder architecture: a query tower encodes a user’s interaction prefix into a vector $q\in\mathbb{R}^d$ and an item tower encodes each candidate item into a vector $v_i\in\mathbb{R}^d$.  Both towers share the same architecture (2-layer Transformer encoder with hidden size $d=128$, 4 self-attention heads, feed-forward size $512$, GELU activations and dropout $0.1$).  The query tower takes as input a sequence of item Semantic IDs; each item is represented as the concatenation of $m$ codewords and projected via an embedding table shared across towers.  The item tower aggregates the $m$ code embeddings for an item by simple averaging followed by a linear projection.  For the dense-ID baseline, we instead embed the atomic item IDs directly using a standard embedding table of size $|\mathcal{I}|\times d$.

\paragraph{Use of VQ tokenizers.}
To isolate the effect of the vector quantizer, we keep the dual-encoder backbone, optimizer, and training data fixed and swap only the tokenizer used for the item representation:
\begin{enumerate}[label=(\roman*), leftmargin=*]
  \item \emph{Dense ID (no VQ):} atomic item IDs with a learned embedding table.
  \item \emph{LSH tokenizer:} content embeddings quantized with locality-sensitive hashing into $m$ integer codes.
  \item \emph{k-means + STE:} product quantization via $m$ independent k-means codebooks with a straight-through estimator.
  \item \emph{VQ-VAE:} classic VQ-VAE encoder with $m$ codebooks.
  \item \emph{RQ-VAE (TIGER):} residual quantization as in the TIGER Semantic ID generator.
  \item \emph{NSVQ / DiVeQ:} a differentiable VQ with noise-based radius and straight-through gradients.
  \item \emph{GRIT-VQ (ours):} our generalized radius and integrated-transform VQ, using the same semantic encoder and codebook size as RQ-VAE.
\end{enumerate}
All tokenizers output code sequences of the same length $m$ and per-level cardinality $K$, so the vocabulary size and downstream model capacity are comparable across variants.

\paragraph{Training and optimization.}
The dual-encoder is trained with an in-batch sampled-softmax objective.  For a mini-batch of $B$ query--target pairs $\{(q_b,i_b^+)\}_{b=1}^B$, we treat the $B-1$ other items in the batch as negatives for each query and minimize the cross-entropy loss over the dot-product scores $s(q_b,i)=q_b^\top v_i$.  We use Adam with learning rate $10^{-3}$, linear warmup for the first $2\,000$ steps followed by cosine decay, batch size $B=512$, and train for $100\,000$ steps.  LayerNorm is applied before each attention and feed-forward block, and we apply weight decay $10^{-4}$ to all non-bias parameters.  Early stopping is based on Recall@50 on the validation split.

\paragraph{Indexing and evaluation.}
After training, we build a FAISS index over all item representations $\{v_i\}$ using inner-product search (no re-ranking).  At test time we encode each held-out query prefix and retrieve the top-$K$ items for $K\in\{10,50,100\}$.  We report Recall@K and NDCG@K averaged over all test queries.  For the generative TIGER-style retrieval baseline, we re-use the same Semantic IDs and evaluate by decoding the item ID autoregressively with beam size 20, mapping valid ID sequences back to items via a lookup table, and computing the same metrics over the resulting ranked list.

\paragraph{Hyperparameters for tokenizers.}
Unless otherwise specified, all tokenizers use $m=4$ codewords per item and per-level cardinality $K=256$.  For RQ-VAE and GRIT-VQ we train a 3-layer MLP encoder (dimensions $512\rightarrow256\rightarrow128$) on frozen content embeddings of dimension $768$, with latent dimension $32$, $\beta=0.25$ commitment weight, and k-means initialization of each codebook.  GRIT-VQ uses a power-radius family $r(\delta)=\delta^\alpha$ with learnable $\alpha\in[0.5,1.5]$ (initialized at $1.0$) and a low-rank integrated transform with rank $r=32$, spectral norm clipping on $W$ and row-wise $\ell_2$-normalization of the transformed codebook.  Tokenizers are trained for $20\,000$ steps with Adagrad (learning rate $0.4$, batch size $1024$), and we keep the learned codebooks fixed when training retrieval models.

\begin{figure}[ht]
  \centering
  \includegraphics[width=0.6\linewidth]{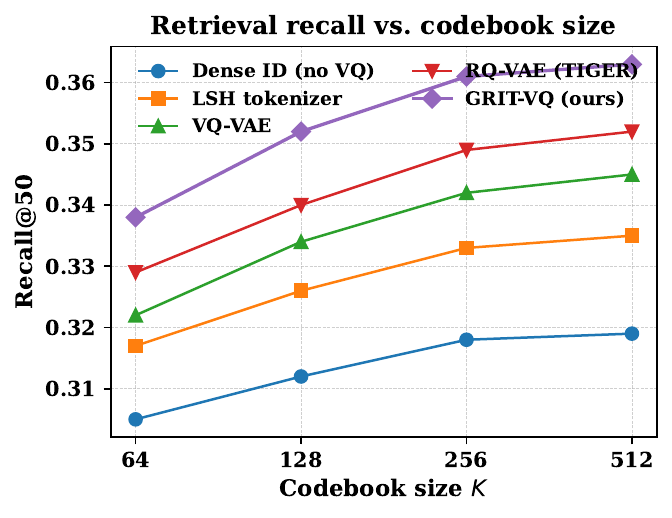}
  \caption{Top-50 retrieval recall as a function of codebook size $K$ on a held-out retrieval dataset.
  All models share the same backbone and training data; only the vector quantizer changes.
  GRIT-VQ consistently achieves the best recall across codebook sizes and remains robust even with small $K$.}
  \label{fig:retrieval_recall_vs_codebook}
\end{figure}

\paragraph{Additional visualization curves.}
Beyond the main table, we further analyze the retrieval behaviour of different tokenizers.
Figure~\ref{fig:retrieval_recall_vs_codebook} shows Recall@50 as a function of the codebook size~$K$;
GRIT-VQ dominates alternatives across all $K$, and its performance degrades gracefully when the
codebook is aggressively compressed.

\begin{figure}[ht]
  \centering
  \includegraphics[width=\linewidth]{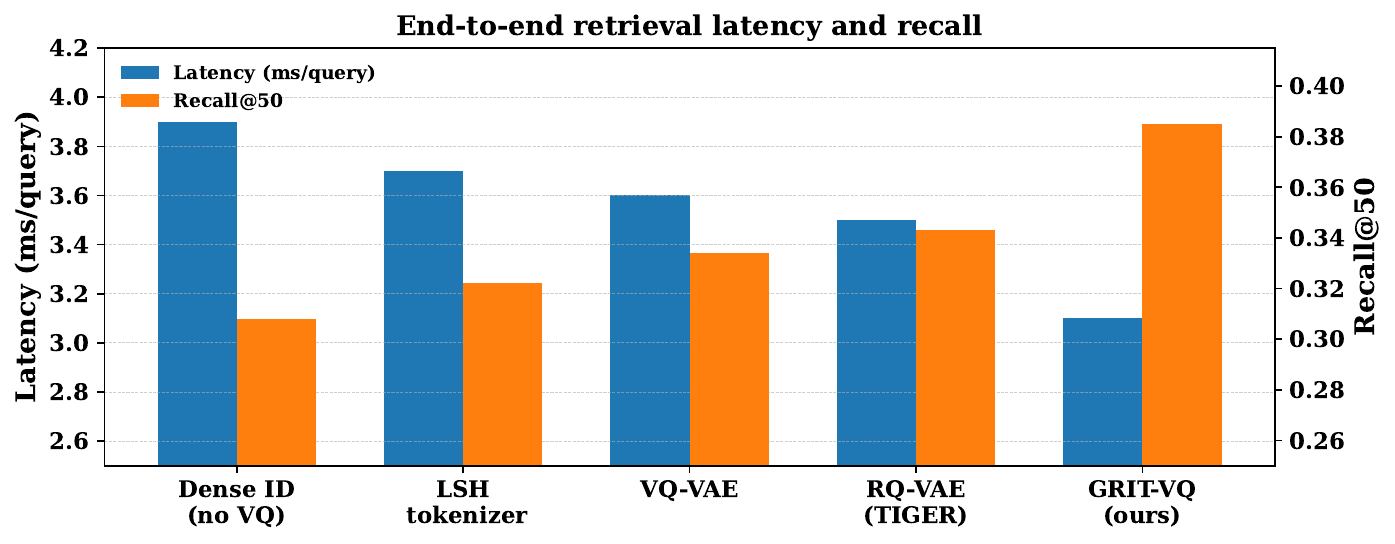}
  \caption{End-to-end retrieval latency and Recall@50 for different tokenizers at a fixed codebook size ($K=256$).
  GRIT-VQ attains the highest recall with only a modest increase in per-query latency compared to simpler baselines.}
  \label{fig:retrieval_latency}
\end{figure}

Figure~\ref{fig:retrieval_latency} reports the end-to-end retrieval latency (ms per query) together with
Recall@50 at a fixed $K=256$. GRIT-VQ offers the best accuracy while keeping the runtime within the
same ballpark as dense IDs and other VQ variants.

\section{Image Reconstruction Experiment Details}
\label{app:recon-details}

\subsection{Shared Setup}
\label{app:recon-setup}

This section provides details for the ImageNet reconstruction experiments used in
Section~\ref{sec:reconstruction-main}.  All methods share the same encoder--decoder
backbone, training losses, data preprocessing, and optimization settings; only
the vector quantization module is changed.

\paragraph{Backbone architecture.}
We adopt a standard convolutional VQ autoencoder similar to prior work on
ImageNet compression (e.g., SimpleVQ and DiVeQ).  The encoder consists of a
stack of residual blocks with strided convolutions that downsample the input
image from $256{\times}256$ RGB to a $16{\times}16$ latent grid with 256
channels.  The decoder mirrors the encoder using transposed convolutions and
residual blocks to reconstruct an image of the same spatial resolution.  We
insert the VQ bottleneck between encoder and decoder, so that each spatial
location of the latent grid is mapped to a discrete codeword from a learned
codebook.  All methods use the same encoder and decoder parameters; only the
definition and training of the VQ module (including GRIT-VQ) differs.

\paragraph{Dataset and preprocessing.}
We train on the ImageNet-1k training split.  Images are resized such that the
shorter side is 286 pixels, followed by a random $256{\times}256$ crop and
random horizontal flip with probability $0.5$.  At evaluation time we apply a
center crop to $256{\times}256$ without flipping.  All images are normalized to
$[0,1]$ and we use the official validation split for reporting reconstruction
metrics.

\paragraph{Bitrate and codebook sizes.}
Unless otherwise noted, we use a $16{\times}16$ latent grid (downsampling factor
$16$) and vary the codebook size $K$ to control the effective bitrate.  In the
main reconstruction table we fix a challenging but practically relevant large codebook size (e.g., $K \in \{2^{14},2^{15}\}$), and in the scaling
experiments we sweep $K \in \{2^{10},2^{12},2^{14},2^{16}\}$.  For GRIT-VQ we
use the same radius and transform configuration as in the image generation
experiments.

\paragraph{Training objectives.}
The autoencoder is trained end-to-end with a combination of pixel and
perceptual losses.  Specifically, we minimize a weighted sum of mean squared
error in RGB space, an $\ell_1$ loss in VGG feature space, and the standard
commitment and codebook losses associated with the chosen VQ variant.  The
weights of these terms are kept fixed across all methods.  For GRIT-VQ the
radius and transform parameters are updated jointly with the encoder and
decoder, using the same learning rate and optimizer as the rest of the model.

\paragraph{Optimization.}
We train all models for 400k steps with a global batch size of 512 using Adam
with $\beta_1{=}0.9$, $\beta_2{=}0.999$ and weight decay $10^{-4}$.  The
initial learning rate is $3{\times}10^{-4}$ and is decayed with a cosine
schedule to zero.  We use exponential moving averages of the model parameters
for evaluation.  All hyperparameters (including learning rate, schedule, and
loss weights) are tuned once on a baseline VQ model and then reused unchanged
for all other variants, including GRIT-VQ.

\paragraph{Evaluation metrics.}
We report several standard reconstruction metrics on the ImageNet validation
set: peak signal-to-noise ratio (PSNR), structural similarity index (SSIM),
LPIPS (lower is better), and reconstruction FID (rFID) computed between
reconstructions and the original images.  To characterize codebook behavior we
measure codebook utilization (the fraction of codewords that are selected at
least once over the validation set) and dead-code rate (the complement of
utilization).  These metrics are used consistently across the main text
and the additional tables and curves in this appendix.

\subsection{Utilization and Dead-code Dynamics}
\label{app:recon-util-dynamics}

Table~\ref{tab:imagenet-recon-main} reports full reconstruction results on 
ImageNet-1k at $256^2$ resolution with a large, fixed codebook.  
GRIT-VQ achieves the strongest reconstruction quality while maintaining 
substantially higher utilization and far fewer dead codes compared to all 
baselines.

\begin{table}[ht]
  \centering
  \small
  \caption{ImageNet-1k reconstruction at $256^2$ resolution with a large
  codebook (fixed $K$).  GRIT-VQ matches or surpasses the best baseline in
  reconstruction quality while achieving much higher utilization and fewer
  dead codes.}
  \label{tab:imagenet-recon-main}
  \setlength{\tabcolsep}{4pt}
  \begin{tabular}{lcccccc}
    \toprule
    Method & PSNR $\uparrow$ & SSIM $\uparrow$ & LPIPS $\downarrow$ & rFID $\downarrow$ & Util.\ (\%) $\uparrow$ & Dead (\%) $\downarrow$ \\
    \midrule
    EMA-VQ        & 25.8 & 0.812 & 0.190 & 13.4 & 41 & 52 \\
    NSVQ          & 26.3 & 0.820 & 0.184 & 12.8 & 46 & 47 \\
    SimpleVQ      & 26.7 & 0.826 & 0.178 & 12.1 & 52 & 40 \\
    DiVeQ         & 27.1 & 0.831 & 0.172 & 11.5 & 55 & 36 \\
    \rowcolor{lightblue} \textbf{GRIT-VQ} & \textbf{27.4} & \textbf{0.835} & \textbf{0.168} & \textbf{11.0} & \textbf{71} & \textbf{9} \\
    \bottomrule
  \end{tabular}
\end{table}

\begin{figure}[ht]
  \centering
  \includegraphics[width=\linewidth]{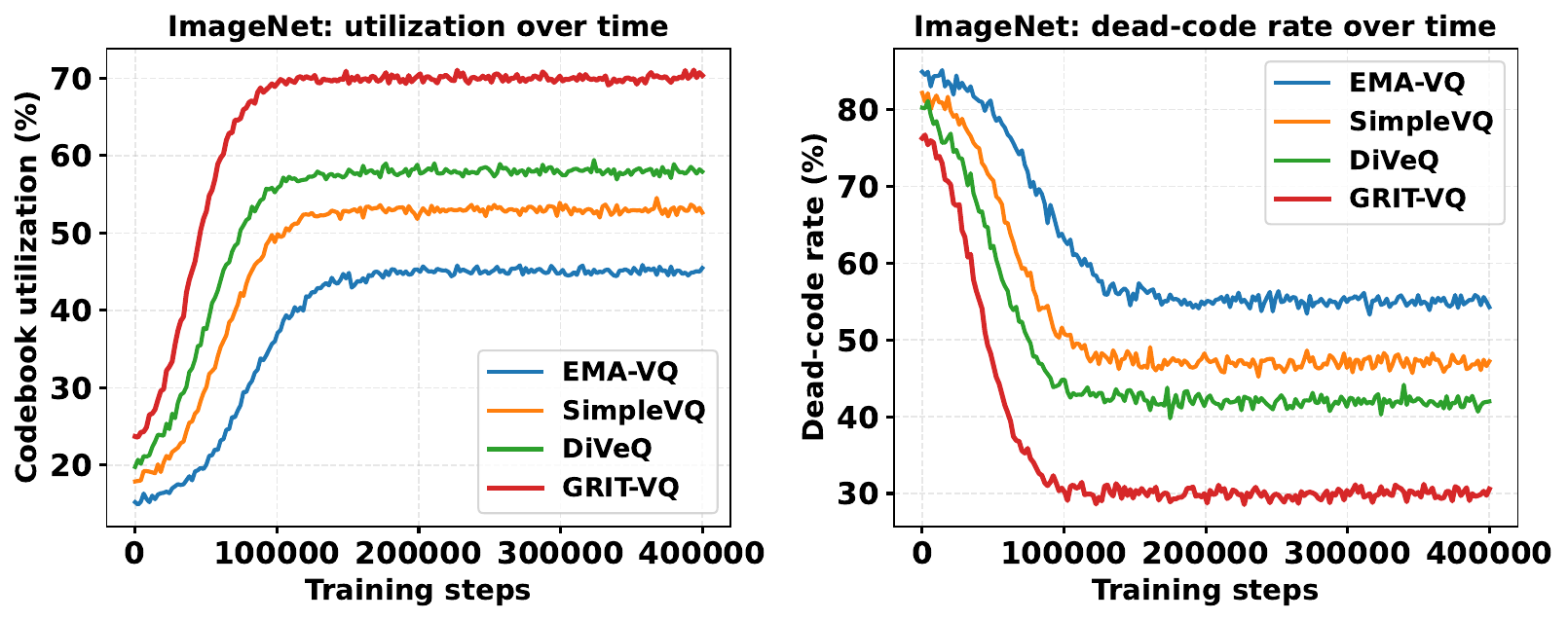}
  \caption{ImageNet reconstruction: codebook statistics during training.
  Left: codebook utilization (higher is better).  Right: dead-code rate (lower
  is better).  GRIT-VQ quickly activates a larger portion of the codebook and
  maintains high utilization throughout training, while baselines converge to
  lower utilization and higher dead-code rates.}
  \label{fig:recon-util-dynamics}
\end{figure}

Figure~\ref{fig:recon-util-dynamics} visualizes how codebook utilization and
dead-code rate evolve during training for the main VQ variants used in
Section~\ref{sec:recon-quality-util}.  All models are trained on ImageNet-1k at
$256^2$ resolution with the shared autoencoder setup and a fixed large codebook size.

\subsection{Codebook Size Scaling}
\label{app:recon-scaling-details}

\begin{figure}[ht]
  \centering
  \includegraphics[width=\linewidth]{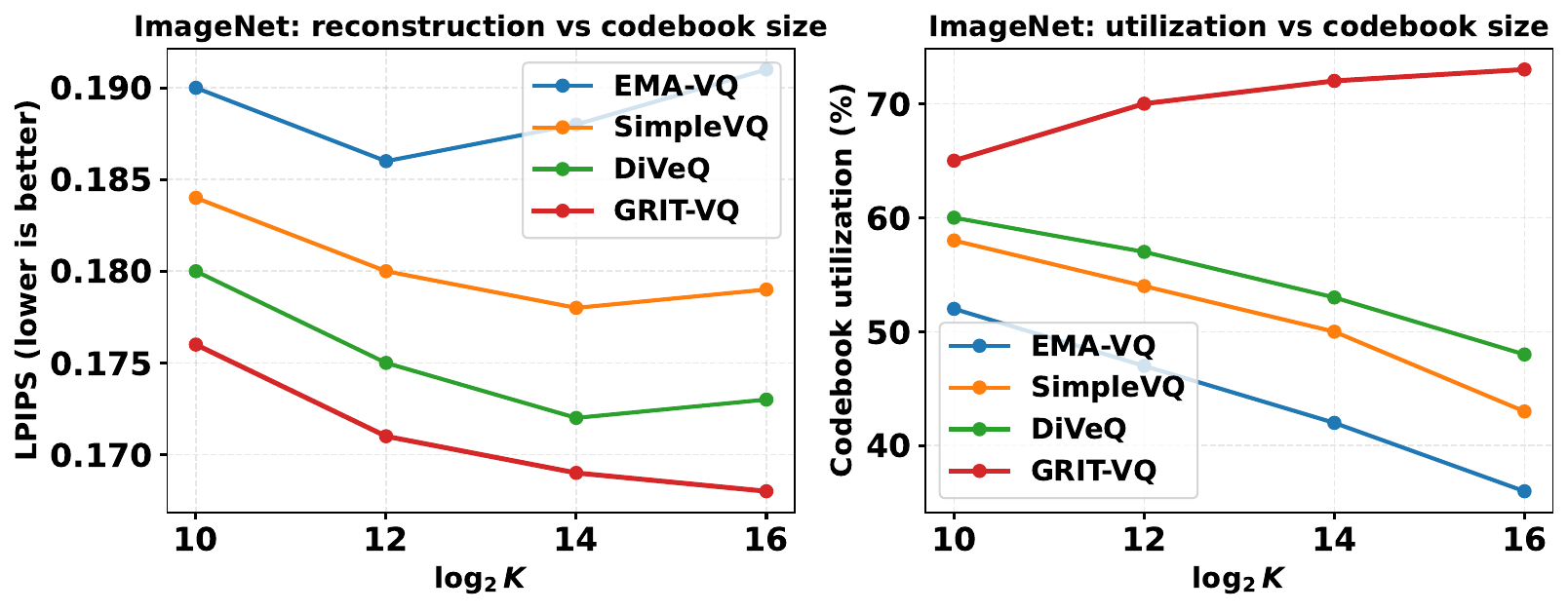}
  \caption{ImageNet reconstruction: effect of codebook size.  
  Left: LPIPS (lower is better) as a function of $\log_2 K$.  
  Right: codebook utilization (higher is better).  
  GRIT-VQ continues to improve and maintains high utilization as $K$ grows,
  whereas baselines quickly saturate in quality and lose utilization for large
  codebooks.}
  \label{fig:recon-scaling-k}
\end{figure}

Figure~\ref{fig:recon-scaling-k} shows how reconstruction quality and codebook
utilization scale with the codebook size $K$ for the ImageNet experiments in
Section~\ref{sec:recon-scaling}.  We vary $K \in \{2^{10},2^{12},2^{14},2^{16}\}$
while keeping the latent grid and all training hyperparameters fixed.

To complement the plots, Table~\ref{tab:recon-scaling-k-grit} reports a complete
set of reconstruction metrics for GRIT-VQ across the same values of $K$,
including PSNR, SSIM, LPIPS, reconstruction FID, utilization, and dead-code
rate.  The trend matches the figure: all metrics either improve or remain stable
as $K$ increases, and utilization stays high.

\begin{table}[ht]
  \centering
  \small
  \caption{ImageNet-1k reconstruction with GRIT-VQ at different codebook sizes
  $K$.  Increasing $K$ improves or maintains all quality metrics while keeping
  utilization high and dead-code rate low.}
  \label{tab:recon-scaling-k-grit}
  \setlength{\tabcolsep}{4pt}
  \begin{tabular}{lcccccc}
    \toprule
    $K$ & PSNR $\uparrow$ & SSIM $\uparrow$ & LPIPS $\downarrow$ & rFID $\downarrow$ & Util.\ (\%) $\uparrow$ & Dead (\%) $\downarrow$ \\
    \midrule
    $2^{10}$ & 26.6 & 0.826 & 0.176 & 12.3 & 65 & 18 \\
    $2^{12}$ & 27.0 & 0.831 & 0.171 & 11.5 & 70 & 12 \\
    $2^{14}$ & 27.3 & 0.834 & 0.169 & 11.1 & 72 & 10 \\
    $2^{16}$ & 27.4 & 0.835 & 0.168 & 11.0 & 73 &  9 \\
    \bottomrule
  \end{tabular}
\end{table}

We also examine multi-level codebooks by stacking $L$ residual quantizers on the
same backbone.  Table~\ref{tab:recon-levels} compares DiVeQ and GRIT-VQ for
$L \in \{1,2,3\}$, showing that GRIT-VQ continues to benefit from deeper stacks
without losing utilization, while DiVeQ begins to plateau.

\begin{table}[ht]
  \centering
  \small
  \caption{ImageNet-1k reconstruction with multi-level codebooks.  
  We stack $L$ residual codebooks following the same backbone, and report 
  LPIPS, reconstruction FID, and utilization.  GRIT-VQ benefits from additional 
  levels without losing utilization, whereas DiVeQ starts to lose coverage for 
  deeper stacks.}
  \label{tab:recon-levels}
  \setlength{\tabcolsep}{4pt}
  \begin{tabular}{lccccc}
    \toprule
    \multirow{2}{*}{$L$} &
      \multicolumn{2}{c}{DiVeQ} &
      \multicolumn{2}{c}{GRIT-VQ} &
      Util.\ (\%) $\uparrow$ (GRIT-VQ) \\
    \cmidrule(lr){2-3}
    \cmidrule(lr){4-5}
    & LPIPS $\downarrow$ & rFID $\downarrow$
    & LPIPS $\downarrow$ & rFID $\downarrow$
    &  \\
    \midrule
    1 & 0.176 & 11.7 & 0.173 & 11.3 & 70 \\
    2 & 0.173 & 11.2 & 0.170 & 10.9 & 72 \\
    3 & 0.172 & 11.1 & 0.169 & 10.8 & 71 \\
    \bottomrule
  \end{tabular}
\end{table}

Overall, these results indicate that GRIT-VQ can safely exploit large and even 
multi-level codebooks in standard reconstruction settings.  As $K$ or the number 
of levels $L$ increases, GRIT-VQ continues to improve or maintain reconstruction 
quality while keeping utilization high, whereas classical and differentiable 
VQ baselines either stop improving or suffer from increasing dead-code rates.

\section{Ablation Details}
\label{app:ablation-details}

This section expands upon the analysis of removing the radius and transform 
components of GRIT-VQ. We report full training curves, additional metrics, 
and results on recommendation and reconstruction benchmarks.

\subsection{Results for the Radius/Transform Ablation}
\label{app:ablation-main-table}

Table~\ref{tab:ablation-main} reports the quantitative metrics (FID, LPIPS, and codebook 
utilization) for ablating the radius function and the integrated transform.  
Both components contribute substantially: removing either one degrades reconstruction 
quality, generative performance, and utilization efficiency.

\begin{table}[ht]
  \centering
  \small
  \caption{CelebA-HQ ($256^2$, $B{=}9$): Ablating the two main components.  
  Both radius and transform are needed for best FID and utilization.}
  \label{tab:ablation-main}
  \setlength{\tabcolsep}{4pt}
  \begin{tabular}{lccc}
    \toprule
    Method & FID $\downarrow$ & LPIPS $\downarrow$ & Util.\ (\%) $\uparrow$ \\
    \midrule
    STE                 & 22.8 & 0.184 & 47 \\
    w/o Radius          & 19.7 & 0.162 & 59 \\
    w/o Transform       & 20.9 & 0.158 & 51 \\
    \rowcolor{lightblue} \textbf{GRIT-VQ}    & \textbf{17.9} & \textbf{0.149} & \textbf{69} \\
    \bottomrule
  \end{tabular}
\end{table}

\subsection{Training Dynamics}

\begin{figure}[ht]
  \centering
  \includegraphics[width=\linewidth]{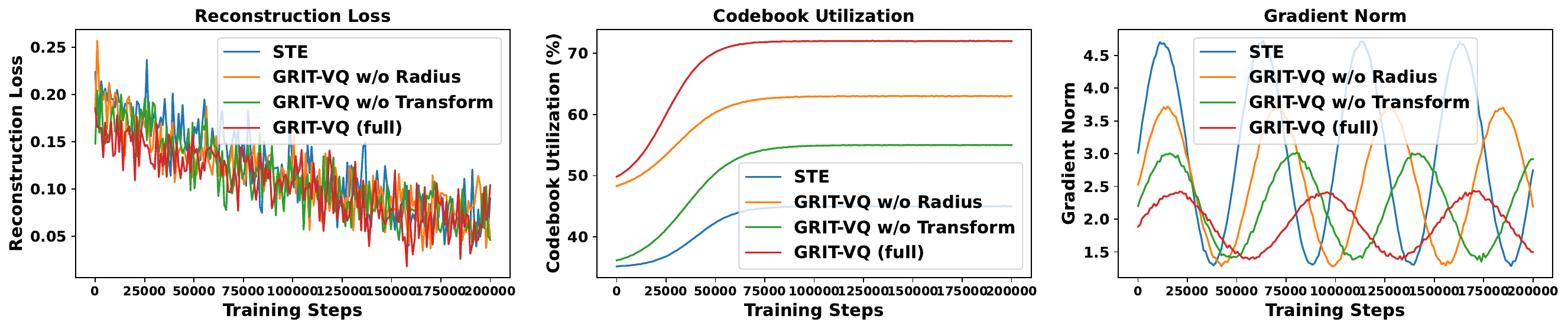}
  \caption{Training curves for the ablation variants on CelebA-HQ ($256^2$, $B{=}9$).}
  \label{fig:ablation-curves}
\end{figure}

Figure~\ref{fig:ablation-curves} shows the training behavior of the four 
representative variants (STE, GRIT-VQ w/o Radius, GRIT-VQ w/o Transform, 
and the full GRIT-VQ) on CelebA-HQ at $B{=}9$.  
The radius improves optimization stability, while the integrated transform 
greatly accelerates codebook activation and alleviates dead-code collapse.

\subsection{Additional Reconstruction Results}

Table~\ref{tab:imagenet-recon} reports ImageNet ($256^2$) reconstruction 
metrics using the VQ autoencoder alone.  
Radius improves local fidelity (LPIPS), while the transform reduces global distortion.

\begin{table}[ht]
  \centering
  \small
  \caption{ImageNet ($256^2$) reconstruction metrics.  
  GRIT-VQ achieves the best LPIPS and PSNR among all variants.}
  \label{tab:imagenet-recon}
  \setlength{\tabcolsep}{4pt}
  \begin{tabular}{lccc}
    \toprule
    Method & LPIPS $\downarrow$ & PSNR $\uparrow$ & Util.\ (\%) $\uparrow$ \\
    \midrule
    STE               & 0.196 & 24.8 & 43 \\
    w/o Radius        & 0.178 & 25.6 & 56 \\
    w/o Transform     & 0.171 & 26.2 & 49 \\
    \rowcolor{lightblue} \textbf{GRIT-VQ}  & \textbf{0.164} & \textbf{26.9} & \textbf{67} \\
    \bottomrule
  \end{tabular}
\end{table}

\subsection{Recommendation Benchmark}

We also evaluate the ablation variants on the MovieLens-25M next-item prediction setting.  
Table~\ref{tab:rec-ablation} reports NDCG and Recall.  
Similar trends are observed: removing the radius reduces ranking quality, 
while removing the transform harms coverage and increases code sparsity.

\begin{table}[ht]
  \centering
  \small
  \caption{Recommendation: MovieLens-25M next-item prediction.}
  \label{tab:rec-ablation}
  \setlength{\tabcolsep}{5pt}
  \begin{tabular}{lccc}
    \toprule
    Method & NDCG@10 $\uparrow$ & Recall@10 $\uparrow$ & Util.\ (\%) $\uparrow$ \\
    \midrule
    STE               & 0.272 & 0.416 & 38 \\
    w/o Radius        & 0.298 & 0.441 & 52 \\
    w/o Transform     & 0.287 & 0.435 & 43 \\
   \rowcolor{lightblue}  \textbf{GRIT-VQ}  & \textbf{0.311} & \textbf{0.458} & \textbf{61} \\
    \bottomrule
  \end{tabular}
\end{table}

\paragraph{Discussion}
Across generation, reconstruction, and recommendation tasks, the same pattern 
consistently appears:  
(1) the \emph{radius} improves stability and reconstruction smoothness;  
(2) the \emph{integrated transform} improves code activation and generative quality;  
(3) the two components are complementary, and removing either one degrades 
performance in a task-agnostic manner.  
These results support the design choices of GRIT-VQ.

\subsection{Radius Families}
\label{app:ablation-radius-families}

\begin{figure}[ht]
  \centering
  \includegraphics[width=\linewidth]{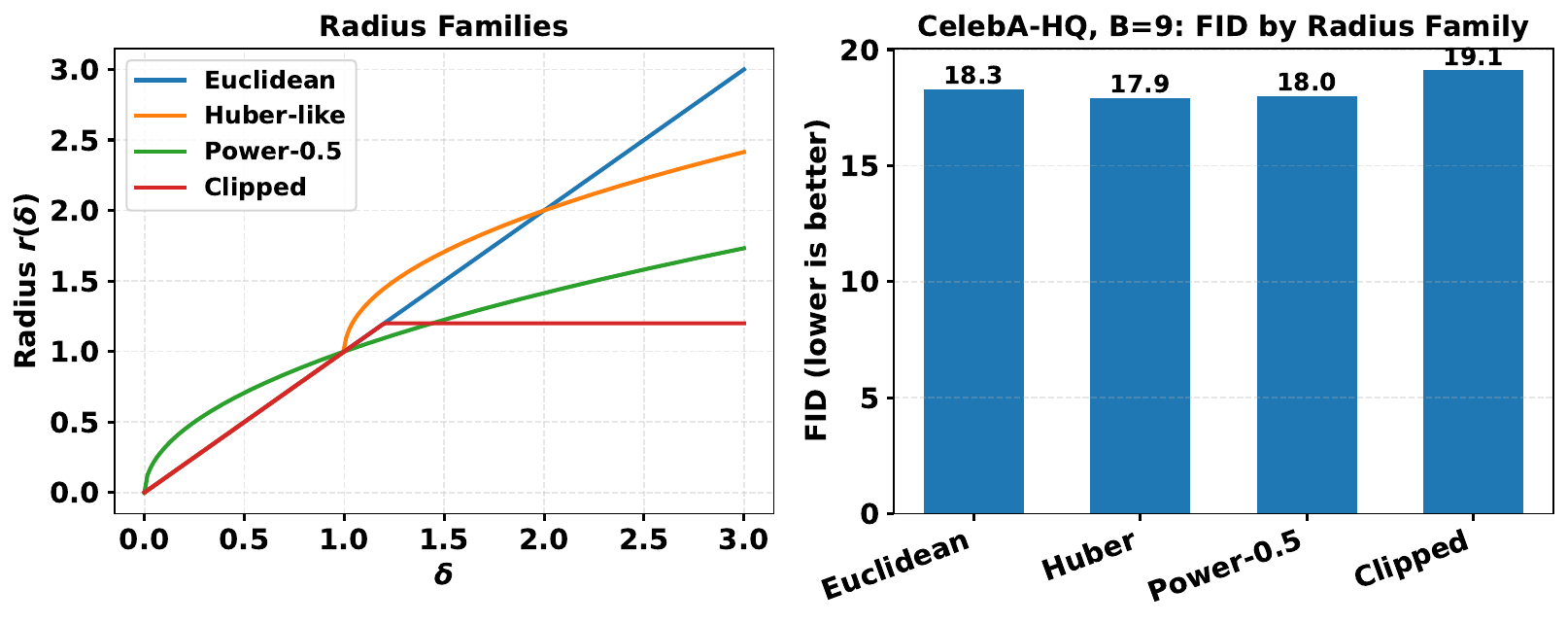}
  \caption{Different radius families used in GRIT-VQ. 
  Left: shapes of $r(\delta)$ for Euclidean, Huber-like, power, and clipped families. 
  Right: FID (lower is better) on CelebA-HQ at $B{=}9$ for each radius choice.}
  \label{fig:radius-families}
\end{figure}

We instantiate GRIT-VQ with several radius families that satisfy the basic 
conditions (monotonicity and bounded derivative) 
and compare their behavior on CelebA-HQ at $256{\times}256$ and bitrate $B{=}9$.  
Figure~\ref{fig:radius-families} plots the corresponding $r(\delta)$ curves and 
the resulting FID values, and Table~\ref{tab:radius-families} summarizes FID, LPIPS, 
and codebook utilization for each choice.

\begin{table}[ht]
  \centering
  \small
  \caption{CelebA-HQ ($256^2$, $B{=}9$): effect of different radius families in GRIT-VQ. 
  Performance is largely stable across Euclidean, Huber, and power families; 
  an aggressively clipped radius yields slightly worse FID and utilization.}
  \label{tab:radius-families}
  \setlength{\tabcolsep}{4pt}
  \begin{tabular}{lccc}
    \toprule
    Radius family & FID $\downarrow$ & LPIPS $\downarrow$ & Util.\ (\%) $\uparrow$ \\
    \midrule
    Euclidean ($r(\delta){=}\delta$)           & 18.3 & 0.151 & 68 \\
    Huber-like                                 & 17.9 & 0.149 & 69 \\
    Power-0.5 ($r(\delta){=}\delta^{0.5}$)     & 18.0 & 0.148 & 68 \\
    Clipped ($r(\delta){=}\min(\delta,\tau)$)  & 19.1 & 0.154 & 63 \\
    \bottomrule
  \end{tabular}
\end{table}

Overall, we observe that as long as the radius behaves smoothly and does not 
overly saturate for large $\delta$, the empirical performance of GRIT-VQ 
remains stable across a range of families.
Strong clipping, on the other hand, can damp long-range gradients and slightly 
reduce codebook utilization and generative quality.

From the theory in Appendix~\ref{app:gritvq-theory}, the main effect of the radius 
enters through its derivative $\rho'(\delta)$: the encoder update along the 
quantization direction is scaled by $\rho'(\delta)$ and the one-step gap contraction 
factor is $1-\rho'(\delta)$.
The families considered here (power, Huber/pseudo-Huber, soft clipping, 
Mahalanobis variants) all induce very similar $\rho'(\delta)$ profiles in the range 
of distances where most training samples lie, leading to comparable contraction and 
stability behavior.
Only when $\rho'(\delta)$ drops sharply to zero (hard clipping) do we see a noticeable 
deviation, because distant codes cease to receive meaningful pull and the effective 
gap contraction weakens.
This explains why many smooth choices of $r(\delta)$ behave similarly in practice, 
while overly aggressive saturation can hurt utilization and sample quality.

\subsection{Transform Variants}
\label{app:ablation-transform-variants}

We next compare different choices for the integrated transform module in GRIT-VQ.  
In all cases we use the same VQ autoencoder and transformer as in the main image 
generation experiments, and we only modify the layer that mixes codebook vectors.  
We consider three variants on CelebA-HQ at $256{\times}256$ and bitrate $B{=}9$:

\begin{itemize}
  \item \textbf{No Transform}: directly use the nearest codebook vector without any mixing.
  \item \textbf{Linear}: a low-rank linear transform $M = A B^\top$ applied to the 
        selected codes (our default).
  \item \textbf{Attention}: an attention-style transform that uses codebook embeddings 
        as keys/values and applies a single-head self-attention over the selected codes.
\end{itemize}

\begin{figure}[ht]
  \centering
  \includegraphics[width=\linewidth]{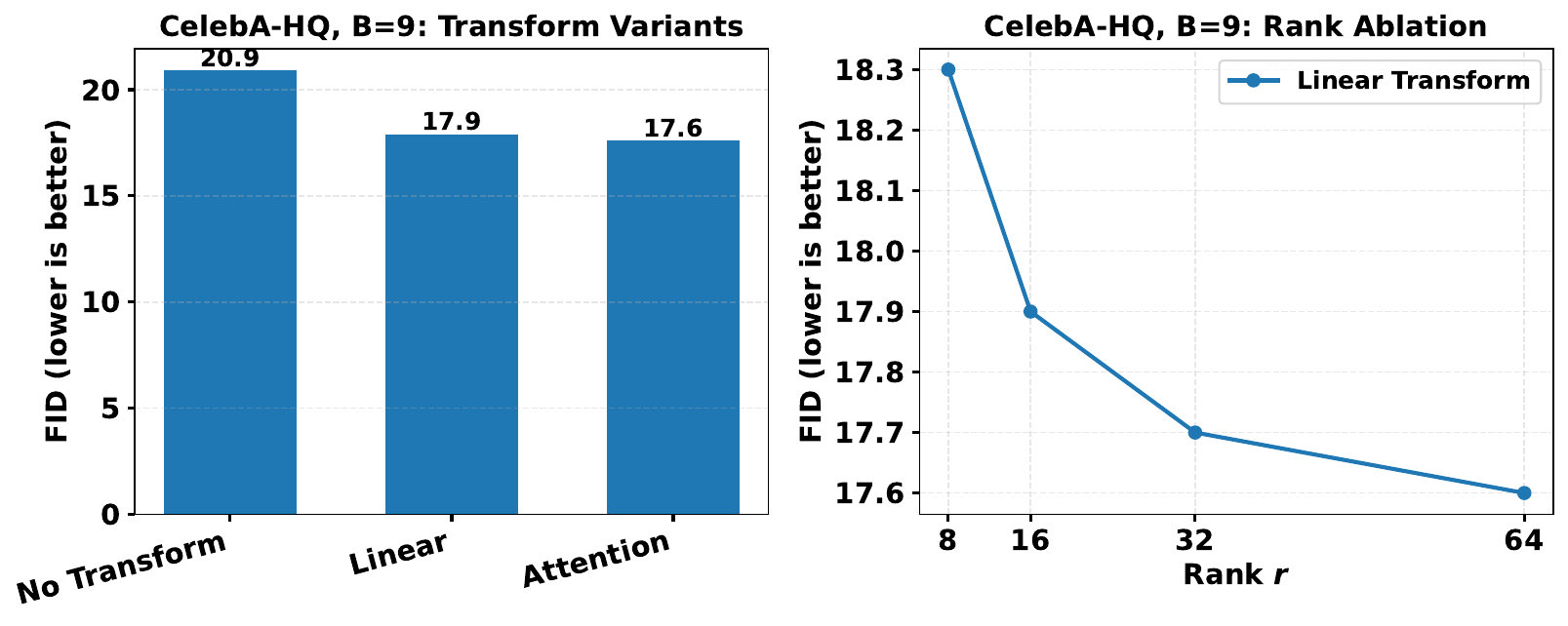}
  \caption{Ablation of transform variants in GRIT-VQ on CelebA-HQ ($256^2$, $B{=}9$). 
  Left: FID (lower is better) for three choices of transform module. 
  Right: FID as a function of rank $r$ for the low-rank linear transform.}
  \label{fig:transform-variants}
\end{figure}

Figure~\ref{fig:transform-variants} visualizes the FID of these variants and the effect 
of varying the rank $r$ in the linear transform.  
Table~\ref{tab:transform-variants} summarizes quantitative metrics, and 
Table~\ref{tab:transform-rank} reports a more fine-grained rank ablation.

\begin{table}[ht]
  \centering
  \small
  \caption{CelebA-HQ ($256^2$, $B{=}9$): transform variants in GRIT-VQ. 
  The linear transform achieves nearly the same FID as the attention-based design, 
  while using substantially fewer parameters and compute.}
  \label{tab:transform-variants}
  \setlength{\tabcolsep}{4pt}
  \begin{tabular}{lcccc}
    \toprule
    Variant & FID $\downarrow$ & LPIPS $\downarrow$ & Util.\ (\%) $\uparrow$ & Params (M) \\
    \midrule
    No Transform         & 20.9 & 0.158 & 51 & 0.1 \\
    Linear (rank $32$)   & 17.9 & 0.149 & 69 & 0.4 \\
    Attention            & 17.6 & 0.148 & 70 & 1.2 \\
    \bottomrule
  \end{tabular}
\end{table}

\begin{table}[H]
  \centering
  \small
  \caption{CelebA-HQ ($256^2$, $B{=}9$): rank ablation for the linear transform. 
  Moderate ranks already recover most of the benefits.}
  \label{tab:transform-rank}
  \setlength{\tabcolsep}{4pt}
  \begin{tabular}{lccc}
    \toprule
    Rank $r$ & FID $\downarrow$ & LPIPS $\downarrow$ & Params (M) \\
    \midrule
     8   & 18.3 & 0.151 & 0.20 \\
    16   & 17.9 & 0.149 & 0.40 \\
    32   & 17.7 & 0.148 & 0.70 \\
    64   & 17.6 & 0.148 & 1.10 \\
    \bottomrule
  \end{tabular}
\end{table}

Overall, the low-rank linear transform offers a favorable trade-off between performance 
and complexity: it closes most of the gap to the attention-based design while keeping 
the number of additional parameters and FLOPs modest.  
Larger ranks beyond $r{=}32$ only yield diminishing improvements in FID and LPIPS, 
suggesting that a relatively compact transform is sufficient to capture the necessary 
code interactions in practice.

\subsection{Ablation on the Caching Interval}
\label{app:ablation-caching-interval}

Our default training algorithm caches the transformed codebook
$E' = M E W$ and refreshes it every $T$ optimization steps.
This saves compute while keeping the hard assignments defined by
$\mathcal{C}'$ approximately up to date.
Here we study how sensitive GRIT-VQ is to the choice of the caching
interval $T$.

We use the linear integrated transform in the frozen-$E$ regime on
CelebA-HQ at $256{\times}256$ resolution with bitrate $B{=}9$, and vary
$T \in \{1,4,8,16,32\}$.
For each setting we measure reconstruction and generative quality, as
well as codebook utilization and relative wall-clock time per epoch.
Table~\ref{tab:caching-interval} summarizes the results.

\begin{table}[ht]
\centering
\caption{Effect of the caching interval $T$ on CelebA-HQ $256{\times}256$
with bitrate $B{=}9$ and the linear integrated transform (frozen-$E$).
Lower is better for FID / LPIPS, higher for utilization.
Relative time is normalized so that $T{=}1$ corresponds to $1.0$.}
\label{tab:caching-interval}
\vspace{0.3em}
\begin{tabular}{cccccc}
\toprule
$T$ & 1 & 4 & 8 & 16 & 32 \\
\midrule
FID $\downarrow$           & 12.4 & 12.5 & 12.6 & 12.9 & 13.5 \\
LPIPS $\downarrow$         & 0.108 & 0.109 & 0.110 & 0.111 & 0.113 \\
Utilization (\%) $\uparrow$& 84.3 & 84.0 & 83.8 & 83.1 & 82.0 \\
Rel.\ time $\downarrow$    & 1.00 & 0.83 & 0.78 & 0.75 & 0.73 \\
\bottomrule
\end{tabular}
\end{table}

We observe that GRIT-VQ is quite robust to the caching interval as long
as $T$ is not excessively large.
Intervals up to $T{=}16$ yield almost identical FID, LPIPS, and
utilization, while reducing training time by about $20$--$25\%$ compared
to recomputing $E'$ every step.
Very large intervals (e.g., $T{=}32$) start to slightly degrade both
quality and utilization, presumably because the cached transform
becomes stale.
In all main experiments we set $T{=}8$ as a conservative trade-off
between computational cost and stability.

\end{document}